\documentclass[ijoc,sglanonrev]{informs4_arxiv}
\usepackage{amsmath}
\RequirePackage{tgtermes}
\RequirePackage{newtxtext}
\RequirePackage{newtxmath}
\RequirePackage{bm}
\RequirePackage{endnotes}

\OneAndAHalfSpacedXII 

\usepackage{amssymb,amsfonts}

\usepackage{natbib}
 \bibpunct[, ]{(}{)}{,}{a}{}{,}%
 \def\bibfont{\small}%

\usepackage{bm}
\usepackage{rotating}
\usepackage{fancyvrb}
\usepackage{tabularx} 
\usepackage{booktabs}
\usepackage{multirow}
\usepackage{caption}
\usepackage{lineno,hyperref}
\usepackage{algorithm}
\usepackage{algorithmicx}
\usepackage{algpseudocode}
\usepackage[T1]{fontenc}
\usepackage{graphicx}
\usepackage{pdflscape}
\usepackage{makecell}
\usepackage{graphicx}
\usepackage{subcaption} 

\makeatletter

\newenvironment{breakablealgorithm}
{
	\begin{center}
		\refstepcounter{algorithm}
		\hrule height.8pt depth0pt \kern2pt
		\renewcommand{\caption}[2][\relax]{
			{\raggedright\textbf{\ALG@name~\thealgorithm} ##2\par}%
			\ifx\relax##1\relax 
			\addcontentsline{loa}{algorithm}{\protect\numberline{\thealgorithm}##2}%
			\else 
			\addcontentsline{loa}{algorithm}{\protect\numberline{\thealgorithm}##1}%
			\fi
			\kern2pt\hrule\kern2pt
		}
	}{
		\kern2pt\hrule\relax
	\end{center}
}
\makeatother
\EquationsNumberedThrough    
\TheoremsNumberedThrough     
\ECRepeatTheorems  %
\MANUSCRIPTNO{IJOC-0001-2024.00}

\def\theAPPENDIXTITLE#1{%
  \HOOKtop
  \begin{flushleft}
  \vspace*{0pt}%
  \TITLEfont\HD{24}{0}\sffamily\bfseries#1\HD{0}{15}%
  \end{flushleft}}

\begin{document}
  \RUNAUTHOR{Wang, Liu, Kadzi{\'n}ski, and Liao}
  
  \RUNTITLE{Preference Construction: A Bayesian Interactive Preference Elicitation Framework Based on Monte Carlo Tree Search}
  
  \TITLE{Preference Construction: A Bayesian Interactive Preference Elicitation Framework Based on Monte Carlo Tree Search}
  
  
  \ARTICLEAUTHORS{%
  \AUTHOR{Yan Wang}
  \AFF{Center for Intelligent Decision-making and Machine Learning, School of Management, Xi'an Jiaotong University, Xi'an 710049, Shaanxi, PR China \\ \EMAIL{wangyan\underline{~}som@stu.xjtu.edu.cn}}
  
  \AUTHOR{Jiapeng Liu\thanks{Corresponding author}}
  \AFF{Center for Intelligent Decision-making and Machine Learning, School of Management, Xi'an Jiaotong University, Xi'an 710049, Shaanxi, PR China \\ \EMAIL{jiapengliu@mail.xjtu.edu.cn}} 
  
  \AUTHOR{Milosz Kadzi{\'n}ski}
  \AFF{Faculty of Computing and Telecommunications, Poznan University of Technology, Piotrowo 2, 60-965 Pozna{\'n}, Poland \\ \EMAIL{milosz.kadzinski@cs.put.poznan.pl}}
  
  \AUTHOR{Xiuwu Liao}
  \AFF{Center for Intelligent Decision-making and Machine Learning, School of Management, Xi'an Jiaotong University, Xi'an 710049, Shaanxi, PR China \\ \EMAIL{liaoxiuwu@mail.xjtu.edu.cn}}
  
  } 
  
  \ABSTRACT{%
  We present a novel preference learning framework to capture participant preferences efficiently within limited interaction rounds. It involves three main contributions. First, we develop a variational Bayesian approach to infer the participant's preference model by estimating posterior distributions and managing uncertainty from limited information. Second, we propose an adaptive questioning policy that maximizes cumulative uncertainty reduction, formulating questioning as a finite Markov decision process and using Monte Carlo Tree Search to prioritize promising question trajectories. By considering long-term effects and leveraging the efficiency of the Bayesian approach, the policy avoids shortsightedness. Third, we apply the framework to Multiple Criteria Decision Aiding, with pairwise comparison as the preference information and an additive value function as the preference model. We integrate the reparameterization trick to address high-variance issues, enhancing robustness and efficiency. Computational studies on real-world and synthetic datasets demonstrate the framework's practical usability, outperforming baselines in capturing preferences and achieving superior uncertainty reduction within limited interactions. 
  }%
  
  
  
  \KEYWORDS{preference learning, decision analysis, preference elicitation, variational Bayesian, Monte Carlo tree search} 
  
  \maketitle
  
  \section{Introduction}\label{sec:Introduction}
  Preferences are ubiquitous and become increasingly important in various fields of application. For example, training Artificial General Intelligence (AGI) models, such as OpenAI's language models, integrates preferences using Reinforcement Learning from Human Feedback (RLHF)~\citep{christiano2017deep} to refine the agent's ability to follow instructions better, maintain factual accuracy, and avoid model hallucinations. Analogously, online streaming platforms, such as Netflix and Spotify, enhance the user experience by leveraging behavior data (including ratings and likes) to analyze implicit preferences and subsequently recommend items that are more likely to engage them~\citep{schedl2021music}. Other relevant applications span social link recommendation (\cite{yin2023Diversity}), computational advertising (\cite{li2024preference}), new product management (\cite{cao2021Preference}), and recommendation systems (\cite{ran2023integrating}). Understanding individual preferences underpins better decision-making.
  
\begin{figure}[!htbp]
	  \centering
	  \includegraphics[scale=0.5]{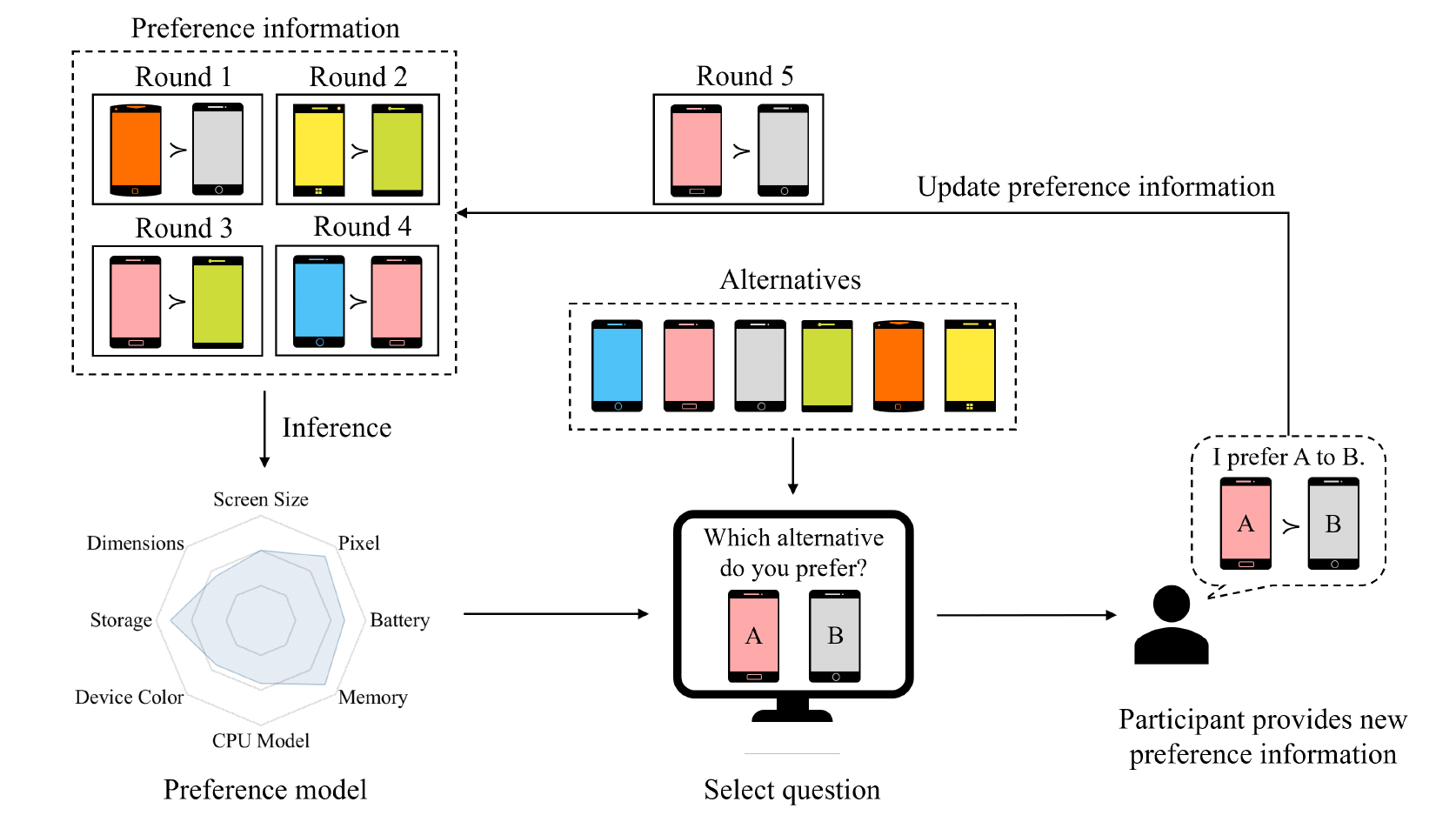}
	  \caption{Preference elicitation process.} \label{fig:introduction}
  \end{figure}

  We focus on interactive preference construction problems, which broadly exist in many fields, such as multiple criteria decision aiding~\citep{corrente2013robust,Greco2024Review}, conjoint analysis in marketing~\citep{dzyabura2011active,dew2024adaptive}, and personalized investment advising in finance~\citep{hu2024interactive}. Such problems can be conceptualized as a sequential preference learning process (see Figure \ref{fig:introduction}). In such an interactive process, participants (e.g., user, decision maker, or stakeholder) incrementally respond to posed questions (e.g., ``Which alternative do you prefer?''). Their answers to these queries (e.g., ``I prefer alternative A to alternative B'') in previous interactions serve as the preference information, which is exploited to guide the subsequent questions for eliciting the participants' further preferences. The ultimate goal of the interactive process is to construct a preference model that can fully characterize their preferences.
  
  The interactive preference construction process must be efficient to prevent participant fatigue and disengagement~\citep{wang2013research, hogan2020neural}. Therefore, the goal is to elicit the most informative preference data within a limited number of interactions. In this regard, two aspects are essential. First, preference information is gathered incrementally, requiring immediate inference of the underlying preference model after each interaction. Early stages face high uncertainty due to limited data, which can hinder the selection of subsequent informative questions if not managed effectively.
  Second, the questioning policy must adapt dynamically based on previous responses, leveraging the inferred preference model to maximize cumulative information gain. The selection and order of questions significantly impact the outcome. Since the process can be modeled as a~decision tree, choosing optimal sub-trees becomes computationally challenging due to factorial growth in possibilities~\citep{ciomek2017heuristics}. Efficient exploration of the search space is thus crucial for effective policy development.
  
  Extensive research highlights the importance of accurate and efficient interactive preference elicitation~\citep{blum2004preference, boutilier2006constraint}. Various approaches have been proposed to address challenges in this area. For preference model inference under uncertainty, one approach constructs models using polyhedral spaces defined by linear constraints from elicited information and employs Monte Carlo simulations to estimate the share of model instances supporting specific preferences~\citep{kadzinski2013robust, ciomek2017heuristics, ciomek2017heuristics2}. While effective for quantifying uncertainty, these methods are vulnerable to cognitive biases in participant responses, potentially excluding the true preference model~\citep{delquie1993inconsistent, ru2022bayesian}. In designing questioning policies, diverse heuristics adaptively select questions to maximize cumulative information gain~\citep{boutilier2006constraint, kadzinski2021active, dzyabura2011active}. However, these heuristics often focus on immediate gains, neglecting long-term impacts and leading to myopic policies. Notable exceptions include \cite{saure2019ellipsoidal}, which uses dynamic programming to propose a one-step look-ahead policy addressing shortsightedness, and \cite{ciomek2017heuristics}, which explores question-answer trees with adjustable search depths. 
  
  This paper introduces a novel Bayesian interactive preference elicitation framework to effectively capture participant preferences within limited interaction rounds. Our major contributions are three-fold.
  First, we develop a variational Bayesian approach to infer preference models. By addressing the preference inference problem from a Bayesian perspective, we derive the posterior distribution of the model from accumulated preference data to manage uncertainty and mitigate cognitive biases. This way, we eliminate the risk of excluding the participant's true preference model. Using variational inference~\citep{murphy2012machine}, we accelerate the computations to under one second, enabling efficient search space exploration for adaptive questioning.
  
  Second, we propose an adaptive questioning policy designed to maximize cumulative uncertainty reduction. Framed as a finite Markov decision process~\citep{sutton2018reinforcement}, it employs Monte Carlo Tree Search (MCTS)~\citep{browne2012survey} to explore the search space efficiently and prioritize promising question trajectories. The introduced policy benefits from the efficient variational Bayesian inference approach. Unlike the existing approaches, it avoids shortsightedness, considering the long-term impacts of decisions by accounting for the states likely to follow and the associated rewards. This way, it selects questions that significantly reduce estimation uncertainty.
  
  Third, we apply the framework to Multiple Criteria Decision Aiding (MCDA), which considers problems with alternatives evaluated on multiple criteria~\citep{corrente2013robust}. Working out a decision recommendation (e.g., ranking of alternatives) requires eliciting preference information from the DM, constructing a well-suited preference model, and applying it to the alternative set~\citep{corrente2013robust}. We use holistic pairwise comparisons as the preference information and an additive value function as the model. In such a setting, the proposed variational Bayesian preference inference approach may encounter high-variance issues that negatively impact the model construction. To address this challenge, we incorporate the reparameterization trick~\citep{kingma2013auto} from machine learning, improving robustness and efficiency in preference inference.
  
  Finally, computational studies validate the framework using real-world and synthetic MCDA datasets. We compare the performance of the proposed variational Bayesian inference approach with or without the reparameterization trick against the baseline approaches. Additionally, we investigate the ability of the proposed questioning policy and baseline heuristics to capture the DM's holistic preferences within a limited number of interaction rounds. 

  The remainder of the paper is organized in the following way. Section \ref{sec:Framework} presents the developed Bayesian interactive preference elicitation framework based on Monte Carlo Tree Search for addressing preference construction problems. Section \ref{sec:Application in MCDA} illustrates the applicability of the proposed framework to the scenario of MCDA. Sections \ref{sec:Experiments on Preference Inference} and \ref{sec:Experiments on Questioning Policy} investigate the performance of the variational Bayesian preference inference approach and the MCTS-based questioning policy through computational experiments involving real-world and artificial datasets with different problem settings, respectively. The last section concludes the paper and provides avenues for future research.

  \section{Bayesian Interactive Preference Elicitation Framework based on Monte Carlo Tree Search}\label{sec:Framework}
  \subsection{Problem Definition and Framework Outline}\label{sec:Problem Definition}
  We consider an interactive preference elicitation problem to comprehensively capture the participant's preferences within a limited number (denoted by $T$) of interaction rounds. 
  In each round, we present the participant with a pair of alternatives ${(a_i, a_j)}$ selected from a set of $n$ alternatives $A=\{a_1,\dots,a_n\}$ for eliciting his/her preferences.
  The participant is allowed to specify a preorder $\succ$ over the pair ${(a_i, a_j)}$ in form of either $a_i \succ a_j$ or $a_j \succ a_i$, representing ``$a_i$ is preferred to $a_j$'' and ``$a_j$ is preferred to $a_i$'', respectively.
  To succinctly illustrate the implementation of the preference elicitation framework, we only consider such strict preference relations (i.e., $a_i \succ a_j$ or $a_j \succ a_i$). 
  However, the framework can be adapted to considering indifference judgments (denoted by $a_i \sim a_j$) along with strict preference relations (see \cite{ru2022bayesian} for more details).
  Let the pairwise comparison provided by the participant in the $t$-th round, $t = 1,\dots,T$, be denoted by $q^{(t)}$. It specifies a preference relation $a_{s_1^{(t)}}\succ a_{s_2^{(t)}}$, such that $s_1^{(t)}, s_2^{(t)} \in \{1,\ldots,n\}$ are the indices of alternatives appearing in this comparison.
  Then, the preference information collected up to $t$-th round can be denoted by $Q^{(t)} = \{q^{(1)},\dots,q^{(t)}\}$.
  
  We postulate the existence of a preference model $U(\cdot)$, which can represent the participant's preferences. 
  The preference model scores the attractiveness of each alternative $a \in A$ as $U(a)$, and the participant is assumed to make pairwise comparisons $q^{(t)}$, $t = 1,\ldots,T$, according to this intrinsic preference model $U(\cdot)$.
  Because the participant's true preference model $U(\cdot)$ is unknown a priori, the objective of the proposed framework is to infer $U(\cdot)$ according to the supplied pairwise comparisons $q^{(t)}$, $t = 1,\ldots,T$.
  In order to infer the preference model $U(\cdot)$ as accurately as possible within $T$ rounds, the proposed framework should address two crucial issues in each round $t$: (a) how to accurately characterize $U(\cdot)$ based on the preference information $Q^{(t)}$ accumulated in the past $t$ rounds? (b) how to select a new pair of alternatives $(a_i, a_j)$ for the participant to make a comparison in the following $(t+1)$-th round, based on the characterization of $U(\cdot)$ derived from the acquired preference information $Q^{(t)}$?
  
  We address these challenges by developing a variational Bayesian approach that infers the participant's preference model $U(\cdot)$ from the preference information $Q^{(t)}$ in each round $t$. We also propose an MCTS-based questioning policy to select a new pair of alternatives $(a_i, a_j)$ for the participant to make a comparison in the following round. The detailed framework, including its phases, is provided in the online supplementary material (Appendix A).
  
  \subsection{Variational Bayesian Approach for Preference Inference}\label{sec:Variational Bayesian Approach}
  
  In this section, we present a variational Bayesian approach for efficiently inferring the participant's preference model $U(\cdot)$ at each round $t$, based on the pairwise comparisons $Q^{(t)}$ accumulated during the previous $t$ rounds of preference elicitation. 
  We address the preference inference problem from a Bayesian perspective, aiming to derive a posterior distribution of the preference model $U(\cdot)$ according to the following Bayes rule:
  
  \begin{small}
  \begin{equation}
	  p(U \mid Q^{(t)}) = \frac{p(Q^{(t)} \mid U)p(U)}{p(Q^{(t)})} = \frac{p(Q^{(t)} \mid U)p(U)}{\int_U p(Q^{(t)} \mid U)p(U) \mathrm{d}U},
  \end{equation}
  \end{small}
  \noindent where $p(U)$ and $p(U \mid Q^{(t)})$ represent the prior and posterior (distributions) of the preference model $U(\cdot)$, respectively, $p(Q^{(t)} \mid U)$ denotes the likelihood of the preference information $Q^{(t)}$ given the preference model $U(\cdot)$, and $p(Q^{(t)}) = \int_U p(Q^{(t)} \mid U)p(U) \mathrm{d}U$ serves as the normalization factor. 
  
  In defining the likelihood $p(Q^{(t)} \mid U)$, we assume the independence between pairwise comparisons in $Q^{(t)}$ and opt for the Bradley-Terry model (\cite{bradley1952rank}) to characterize the probability of generating each pairwise comparison $q^{(d)}$, $d=1,\ldots,t$, by the underlying preference model $U(\cdot)$. In this way, the likelihood $p(Q^{(t)} \mid U)$ can be formulated as:
  
  \begin{small}
	  \begin{equation}
		  p(Q^{(t)} \mid U) =\prod_{d=1}^t p(q^{(d)}: a_{s_1^{(d)}}\succ a_{s_2^{(d)}} \mid U)
		  =\prod_{d=1}^t \frac{\exp(U(a_{s_1^{(d)}}))}{\exp(U(a_{s_1^{(d)}}))+\exp(U(a_{s_2^{(d)}}))}.
	  \end{equation}\end{small}
  This formulation has been adopted in \cite{ru2022bayesian}, in which one can find more details for justifying this formulation.
  
  The prior $p(U)$ specifies the distribution of the preference model $U(\cdot)$ before any preference information is provided by the participant.
  For simplicity, we assume $p(U)$ is a uniform distribution on the set of all potential preference model instances.
  One can refer to \cite{ru2022bayesian} for discussion on specifying a customized prior $p(U)$ by incorporating the participant's subjective judgments about the decision problem.
  
  Once the likelihood $p(Q^{(t)} \mid U)$ and the prior $p(U)$ have been defined, our attention turns to deriving the posterior:
  \begin{small}
  \begin{equation}
	  p(U \mid Q^{(t)}) = \frac{\prod_{d=1}^t \frac{\exp(U(a_{s_1^{(d)}}))}{\exp(U(a_{s_1^{(d)}}))+\exp(U(a_{s_2^{(d)}}))}  p(U)}{\int_{U} \prod_{d=1}^t \frac{\exp(U(a_{s_1^{(d)}}))}{\exp(U(a_{s_1^{(d)}}))+\exp(U(a_{s_2^{(d)}}))} p(U) \mathrm{d}U}.
  \end{equation}
  \end{small}
  The posterior $p(U \mid Q^{(t)})$ assigns each potential preference model instance $U(\cdot)$ a probability mass that signifies its fitting ability to reproduce the preference information $Q^{(t)}$. 
  Following the principle of Bayes rule, the posterior $p(U \mid Q^{(t)})$ strikes a balance between the likelihood $p(Q^{(t)} \mid U)$ and the prior $p(U)$. 
  When the pairwise comparisons in $Q^{(t)}$ are few (i.e., $t$ is small), the impact of the likelihood $p(Q^{(t)} \mid U)$ on the posterior $p(U \mid Q^{(t)})$ is marginal and the posterior $p(U \mid Q^{(t)})$ tends to be uniformly distributed on the set of potential preference model instances (recall that the prior $p(U)$ is a uniform distribution).
  As the number of pairwise comparisons in $Q^{(t)}$ increases (i.e., $t$ rises), the likelihood $p(Q^{(t)} \mid U)$ overwhelms the prior $p(U)$, leading the posterior $p(U \mid Q^{(t)})$ to converge closer to the participant's true preference model instance (\cite{ru2022bayesian}).
  
  It is beneficial to address the preference inference problem within the Bayesian framework due to the following two reasons. 
  First, the Bayesian preference inference method represents the participant's preferences using the derived posterior $p(U \mid Q^{(t)})$ defined over all the potential preference model instances rather than arbitrarily selecting a single preference model instance. 
  Taking into account the whole set of potential preference model instances would benefit the following selection of the next pair of alternatives for inquiry (see Section \ref{sec:Questioning Policy}) since the uncertainty in exploring the preference information $Q^{(t)}$ is fully addressed within the Bayesian framework and implied by the posterior $p(U \mid Q^{(t)})$. 
  Second, the Bayesian preference inference method tolerates the participant's cognitive biases underlying the preference information $Q^{(t)}$ by assigning probability masses to all the potential preference model instances, not necessarily only the compatible ones which reconstruct all the pairwise comparisons in $Q^{(t)}$. 
  At this point, the Bayesian preference inference method is robust in characterizing the participant's preferences, as any potential cognitive bias cannot exclude the true preference model instance from the consideration set.
  
  While conceptually straightforward, the practical implementation of this Bayesian preference inference method confronts a challenge that the posterior $p(U \mid Q^{(t)})$ cannot be derived in an analytical form.
  This is because the prior $p(U)$ is not conjugate to the likelihood $p(Q^{(t)} \mid U)$, which results in no closed-form solution for the normalization factor $p(Q^{(t)}) = \int_U p(Q^{(t)} \mid U)p(U) \mathrm{d}U$ (\cite{rossi2012bayesian}).
  Therefore, it is necessary to resort to using an effective inference technique to approximate the target distribution $p(U \mid Q^{(t)})$.
  The approximation inference methods for Bayesian models in the literature can be broadly categorized into sampling- and optimization-based (see \cite{murphy2012machine, murphy2023probabilistic}). 
  Typical instances of the sampling-based approximation inference methods include Markov Chain Monte Carlo (MCMC) methods and their variants.
  These approaches iteratively sample parameter values from the full conditional distributions and perform the inference based on the sample of correlated draws.
  The drawback of such methods consists in their low efficiency in data-intensive tasks due to the need for tens of thousands
  of iterations to achieve convergence.
  By contrast, variational Bayes (also called variational inference) methods, as representatives of the optimization-based inference methods, replace sampling with optimization and approximate the posterior distribution with a variational distribution by solving an optimization problem. These methods enhance the speed and scalability of model inference and, thus, are especially suitable for large-scale applications.
  In light of this advantage, variational Bayes can provide efficient model inference for our scenario because the proposed MCTS-based search method for selecting a new inquiry question $(a_i ? a_j)$ involves numerous model inference problem instances, and thus the efficiency of the inference method is the utmost concern in this paper (see Section \ref{sec:Questioning Policy}).
  
  We give an approximation inference approach based on variational Bayes for approximating the posterior $p(U \mid Q^{(t)})$ of the preference model $U(\cdot)$.
  Assume the preference model $U(\cdot)$ is parameterized by a parameter vector $\bm{u}$ and its prior is specified by another parameter vector $\bm{\alpha}$. 
  Then, the prior $p(U)$, the likelihood $p(Q^{(t)} \mid U)$, and the posterior $p(U \mid Q^{(t)})$ can be, respectively, rewritten as $p(\bm{u} \mid \bm{\alpha})$, $p(Q^{(t)} \mid \bm{u})$, and $p(\bm{u} \mid Q^{(t)})$.
  The proposed approximation inference approach approximates the posterior $p(\bm{u} \mid Q^{(t)})$ using a simple variational distribution $q(\bm{u} \mid \bm{\theta})$ parameterized by a variational parameter vector $\bm{\theta}$.
  The difference between $p(\bm{u} \mid Q^{(t)})$ and $q(\bm{u} \mid \bm{\theta})$ can be quantified using the Kullback-Leibler divergence defined as $D_{\mathbb{KL}}\left(q\left(\bm{u} \mid \bm{\theta}\right) \mid\mid p\left(\bm{u} \mid Q^{\left(t\right)}\right)\right) = \int q\left(\bm{u} \mid \bm{\theta}\right) \log \frac{q\left(\bm{u} \mid \bm{\theta}\right)}{p\left(\bm u \mid Q^{\left(t\right)}\right)} \mathrm{d}\bm{u}$.
  Variational Bayes aims to minimizes the KL divergence by adjusting $\bm{\theta}$ to find the best approximation $q(\bm{u} \mid \bm{\theta})$ to $p(\bm{u} \mid Q^{(t)})$.
  However, the KL divergence cannot be optimized directly with respect to $\bm{\theta}$.
  Instead, minimizing the KL divergence can be equivalently transformed to maximizing the evidence lower bound (ELBO) function $L(\bm{\theta})$ with respect to $\bm{\theta}$ as follows (\cite{murphy2023probabilistic}):
  
  \begin{small}
  \begin{align}
  \mathop {\max}\limits_{\bm{\theta}} L(\bm{\theta})  = \int q(\bm{u}\mid \bm{\theta})\log \frac{p(Q^{(t)}, \bm u \mid \bm{\alpha})}{q(\bm{u} \mid \bm{\theta})} \mathrm{d}\bm{u}.
  \end{align}
  \end{small}
  
  The above problem can be addressed by applying gradient-based optimization algorithms.
  However, the gradient of the ELBO function $L(\bm\theta)$ with respect to the variational parameter vector $\bm{\theta}$, denoted by $\nabla_{\bm{\theta}}L(\bm{\theta})$, cannot be computed analytically, due to the lack of conjugacy between the prior $p(\bm{u} \mid \bm{\alpha})$ and the likelihood $p(Q^{(t)} \mid \bm{u})$.
  Instead, the gradient $\nabla_{\bm{\theta}}L(\bm{\theta})$ can be approximated using Monte Carlo sampling as follows (see Online Appendix B):
  
  \begin{small}
  \begin{equation} \label{eq:5}
	  \nabla_{\bm{\theta}}L(\bm{\theta})
	  \approx\sum\limits_{w=1}^{W} [\nabla_{\bm\theta}\log q(\bm{u}^{(w)} \mid \bm\theta)][\log p(Q^{(t)} \mid \bm{u}^{(w)}) + \log p(\bm{u}^{(w)} \mid \bm{\alpha})-\log q(\bm{u}^{(w)} \mid \bm\theta)],
  \end{equation}
  \end{small}
  
  \noindent where $\bm{u}^{(w)}$, $w=1,\ldots,W$, are $W$ samples of the parameter $\bm{u}$ drawn from the distribution $q(\bm{u} \mid \bm\theta)$.
  Then, gradient-based optimization algorithms can be used based on the derived stochastic gradient $\nabla_{\bm{\theta}}L(\bm{\theta})$ to maximize the ELBO function with respect to $\bm{\theta}$.
  Finally, the variational distribution $q(\bm{u} \mid \bm{\theta}^*(Q^{(t)}))$ equipped with the achieved optimum for $\bm{\theta}$, denoted by $\bm{\theta}^*(Q^{(t)})$, is the best approximation to the posterior $p(\bm{u} \mid Q^{(t)})$.
  Note that the optimum $\bm{\theta}^*(Q^{(t)})$ depends on the preference information $Q^{(t)}$, meaning that the optimum variational distribution varies at different rounds.

  \subsection{MCTS-based Questioning Policy for Question Selection}\label{sec:Questioning Policy}
  
  This section introduces a questioning policy for the interactive preference elicitation problem. 
  The questioning policy aims to maximize the accumulated variance reduction scales in estimating the participant's preference model.
  At each round $t=1,\ldots,T$, the MCTS-based questioning policy iteratively selects a new pair of alternatives $(a_i, a_j)$ for the participant to make a comparison conditional on the current accumulated preference information $Q^{(t-1)}$.
	  
  First, let us consider quantifying the value of a question $(a_i ? a_j)$. 
  At the end of the $(t-1)$-th round, the accumulated preference information is $Q^{(t-1)}$.
  Assume the participant's preferences are stable during the whole procedure, and then we can predict the participant's answer to a new question $(a_i ? a_j)$ at the $t$-th round based on $Q^{(t-1)}$ as follows.
  The probability of the participant expressing $a_i \succ a_j$ is defined as the following posterior predictive probability:
  
  \begin{small}
  \begin{equation}
	  \begin{aligned}
		  p(a_i \succ a_j \mid Q^{(t-1)}) &= \int p(a_i \succ a_j, U \mid Q^{(t-1)}) \mathrm{d}U\\
		  &= \int p(a_i \succ a_j \mid U) p(U \mid Q^{(t-1)}) \mathrm{d}U\\
		  &=\int p(a_i \succ a_j \mid \bm{u}) p(\bm{u} \mid Q^{(t-1)})\mathrm{d}\bm{u}\\
		  &= \mathbb{E}_{\bm{u} \sim p(\bm{u} \mid Q^{(t-1)})} p(a_i \succ a_j \mid \bm{u}).
	  \end{aligned}
  \end{equation}
  \end{small}
  
  Analogously, the probability of the participant expressing $a_j \succ a_i$ is defined as:
  \begin{small}
  \begin{equation}
  \begin{aligned}
  p(a_j \succ a_i \mid Q^{(t-1)}) &= \mathbb{E}_{\bm{u} \sim p(\bm{u} \mid Q^{(t-1)})} p(a_j \succ a_i \mid \bm{u}).
  \end{aligned}
  \end{equation}
  \end{small}
  Since the posterior $p(\bm{u} \mid Q^{(t-1)})$ is approximated by the variational distribution $q(\bm{u} \mid \bm{\theta}^*(Q^{(t-1)}))$,  the above posterior predictive distribution can be calculated as:
  \begin{small}
  \begin{equation}
		  p(a_i \succ a_j \mid Q^{(t-1)}) \approx \mathbb{E}_{\bm{u} \sim q(\bm{u} \mid \bm{\theta}^*(Q^{(t-1)}))} p(a_i \succ a_j \mid \bm{u})
		  \approx \frac{1}{W} \sum\limits_{w=1}^{W} \mathbb{I}(U^{(w)}(a_i) > U^{(w)}(a_j)),
  \end{equation}
  \end{small}
  \begin{small}    
  \begin{equation}
		  p(a_j \succ a_i \mid Q^{(t-1)}) \approx \mathbb{E}_{\bm{u} \sim q(\bm{u} \mid \bm{\theta}^*(Q^{(t-1)}))} p(a_j \succ a_i \mid \bm{u})
		  \approx \frac{1}{W} \sum\limits_{w=1}^{W} \mathbb{I}(U^{(w)}(a_j) > U^{(w)}(a_i)),
  \end{equation} 
  \end{small}  
  \noindent where $W$ preference model instances $U^{(1)}(\cdot), \ldots, U^{(W)}(\cdot)$ are equipped with $W$ samples of the parameter $\bm{u}$ drawn from the posterior $q(\bm{u} \mid \bm{\theta}^*(Q^{(t-1)}))$ to indicate whether $U^{(w)}(a_i) > U^{(w)}(a_j)$ or $U^{(w)}(a_j) > U^{(w)}(a_i)$, $w=1,\ldots,W$. The function $\mathbb{I}(U^{(w)}(a_i) > U^{(w)}(a_j))$ (or $\mathbb{I}(U^{(w)}(a_j) > U^{(w)}(a_i))$) is an indicator such that $\mathbb{I}(U^{(w)}(a_i) > U^{(w)}(a_j)) = 1$ (or $\mathbb{I}(U^{(w)}(a_j) > U^{(w)}(a_i)) = 1$) if $U^{(w)}(a_i) > U^{(w)}(a_j)$ (or $U^{(w)}(a_j) > U^{(w)}(a_i)$) and 0, otherwise. It is obvious that $p(a_i \succ a_j \mid Q^{(t-1)}) + p(a_j \succ a_i \mid Q^{(t-1)}) = 1$.
  
  With the posterior predictive probabilities $p(a_i \succ a_j \mid Q^{(t-1)})$ and $p(a_j \succ a_i \mid Q^{(t-1)})$ defined above, we can quantify the value of a question $(a_i ? a_j)$ by measuring the uncertainty reduction in exploring the participant's preferences, as we aim to capture the participant's preferences comprehensively.
  Consequently, the value of a question $(a_i ? a_j)$ at the $t$-th round (denoted by $v^{(t)}(a_i, a_j)$) can be defined as the expected scale of variance reduction if we inquire about the participant's preference between $a_i$ and $a_j$, i.e.,
  
  \begin{small}
  \begin{equation} \label{eq:10}
	  \begin{aligned}
	  v^{(t)}(a_i, a_j)=
	  & p(a_i \succ a_j \mid Q^{(t-1)})
	  \{\text{Var}[p(\bm{u} \mid Q^{(t-1)})]-\text{Var}[p(\bm{u} \mid Q^{(t-1)} \cup \{a_i \succ a_j\})]\} +\\
	  & p(a_j \succ a_i \mid Q^{(t-1)})
	  \{\text{Var}[p(\bm{u} \mid Q^{(t-1)})]-\text{Var}[p(\bm{u} \mid Q^{(t-1)} \cup \{a_j \succ a_i\})]\} \\	
	  \approx & p(a_i \succ a_j \mid Q^{(t-1)})
	  \{\text{Var}[p(\bm{u} \mid \bm{\theta}^*(Q^{(t-1)}))]-\text{Var}[p(\bm{u} \mid \bm{\theta}^*(Q^{(t-1)} \cup \{a_i \succ a_j\}))]\} +\\
	  & p(a_j \succ a_i \mid Q^{(t-1)})
	  \{\text{Var}[p(\bm{u} \mid \bm{\theta}^*(Q^{(t-1)}))]-\text{Var}[p(\bm{u} \mid \bm{\theta}^*(Q^{(t-1)} \cup \{a_i \succ a_j\}))]\}, \\	
	  \end{aligned}
  \end{equation}
  \end{small}
  
  \noindent where \( q(\bm{u} \mid \bm{\theta}^*(Q^{(t-1)} \cup \{a_i \succ a_j\})) \) and \( q(\bm{u} \mid \bm{\theta}^*(Q^{(t-1)} \cup \{a_j \succ a_i\})) \) are the optimum variational distributions for approximating the posteriors \( p(\bm{u} \mid Q^{(t-1)} \cup \{a_i \succ a_j\}) \) and \( p(\bm{u} \mid Q^{(t-1)} \cup \{a_j \succ a_i\}) \), respectively, given the prior observations \( Q^{(t-1)} \) and the newly observed preference. Such a definition considers the uncertainty of whether the participant prefers $a_i$ to $a_j$ or $a_j$ to $a_i$.
  One can see that the value $v^{(t)}(a_i, a_j)$ of a question $(a_i ? a_j)$ varies depending on the timing of an inquiry, i.e., at which round $t$ the participant expresses her preference between $a_i$ and $a_j$.
  
  Based on the quantification of the value of a question during the procedure, the goal of the questioning policy can be formulated to maximize the accumulated values of the questions inquired to the participant within the $T$ rounds.
  Such a problem can be considered as a finite Markov decision process (\cite{sutton2018reinforcement}), in which the accumulated preference information $Q^{(t-1)}$ is the state at the $t$-th round, the posed question $(a_i ? a_j)$ is the action, and the uncertainty reduction of the posterior of the participant's preference model is the reward.
  In particular, the explicit representations of the transition and reward models of this problem can be exactly built.
  Specifically, after posing a question \( (a_i ? a_j) \) at the \( t \)-th round, the new state \( Q^{(t)} \) is given by \( Q^{(t-1)} \cup \{a_i \succ a_j\} \) with probability \( p(a_i \succ a_j \mid Q^{(t-1)}) \) or \( Q^{(t-1)} \cup \{a_j \succ a_i\} \) with probability \( p(a_j \succ a_i \mid Q^{(t-1)}) \). The corresponding reward is defined as the reduction in posterior variance:  
  $\text{Var}[p(\bm{u} \mid Q^{(t-1)})] - \text{Var}[p(\bm{u} \mid Q^{(t-1)} \cup \{a_i \succ a_j\})]$ or $\text{Var}[p(\bm{u} \mid Q^{(t-1)})] - \text{Var}[p(\bm{u} \mid Q^{(t-1)} \cup \{a_j \succ a_i\})].$ 
  Therefore, to address this problem, we can use model-based reinforcement learning methods, which use a planning algorithm based on the constructed transition and reward models to form the questioning policy (\cite{murphy2023probabilistic}).
  Dynamic programming is a classical representative of this stream, but it is of low efficiency in dealing with the considered problem.
  This is because dynamic programming needs to perform sweeps through the entire state-action space and update each state-action pair once per sweep to generate the distribution of possible transitions for each state. 
  Then, each distribution is used to compute a backed-up value (update target) and update the state's estimated value. 
  Such a way is problematic in a large state-action space as there may be insufficient computational budget to complete even one sweep (\cite{sutton2018reinforcement}). 
  The planning algorithm used in the way, e.g., dynamic programming, is called \emph{background planning}.
  The other way to use planning in Markov decision processes, named \emph{decision-time planning}, is to begin and complete it after observing each new state as a computation whose output selects a single action.
  In contrast to background planning, decision-time planning focuses more on the current state by dedicating the limited computational budget to make rollout to simulate sequences that all begin at the current state.
  Decision-time planning is more suitable for the problem considered in this paper because the questioning policy in the interaction with the participant should be more specific to the current state rather than exploring the entire state-action space before beginning the interactive preference elicitation process.
  
  This paper applies a state-of-the-art decision-time planning algorithm called Monte Carlo Tree Search (MCTS) to propose a questioning policy for the interactive preference elicitation problem.
  MCTS is mainly responsible for the success of the famous AlphaGo and AlphaZero programs (see \cite{silver2016mastering}) and has been widely applied to many other kinds of sequential decision problems (see \cite{browne2012survey}).
  The core idea of MCTS is to successively focus multiple simulations starting at the current state by extending the initial portions of sequences that have received high evaluations from earlier simulations (\cite{sutton2018reinforcement}). 
  This feature can mitigate short-sightedness by considering potential future occurrences while identifying valuable questions through a certain mechanism and facilitating random exploration to some extent, thus alleviating the computational burden associated with fully exploring the state-action space in terms of computation time or space. 
  
  We provide an overview of the implementation of the MCTS-based questioning policy, while a detailed description, including algorithmic formulation and theoretical foundations, is presented in the Online Appendix C due to space constraints.
  Let $\pi(\cdot)$ denote the questioning policy that selects the pairwise comparison $(a_i ? a_j)$ (action) to present at each round $t = 1, \dots, T$, based on the accumulated preference information $Q^{(t-1)}$ (state).
  At each interaction round $t$, the questioning policy $\pi(Q^{(t-1)})$ executes an MCTS procedure, performing as many iterations as possible within the predefined computational budget before selecting the next question for the participant. This process incrementally constructs a question tree, where the root node represents the current state, and the child nodes correspond to candidate questions.
  Each iteration consists of four key operations: \emph{Selection}, \emph{Expansion}, \emph{Simulation}, and \emph{Backpropagation}. Finally, once the predefined computational budget is reached, the root's child with the highest averaged reward will be chosen to indicate the corresponding question $\left(a_i ? a_j\right)$ to be posed in the $t$-th round.
  
  \section{Application of Bayesian Interactive Preference Elicitation Framework in MCDA}\label{sec:Application in MCDA}
  In this section, we expound upon applying the proposed Bayesian interactive preference elicitation framework in the Multiple Criteria Decision Aiding (MCDA) scenario. 
  MCDA aims to assist the DM in assessing a finite set of alternatives $A=\{a_1, \ldots, a_n\}$ in terms of multiple criteria $G=\{g_1, \ldots, g_m\}$. 
  Let $g_j(a_i)$ denote the performance of alternative $a_i \in A$ on criterion $g_j \in G$. 
  Without loss of generality, we assume that the larger $g_j(a_i)$, the more preferred alternative $a_i$ on criterion $g_j$.
  In the context of MCDA, the most widely used type of preference model to represent the DM's preferences is the additive value function model defined as follows (\cite{keeney1993decisions}):
  
  \begin{small}
  \begin{equation}
	  U(a_i)=\sum_{j=1}^m u_j(g_j(a_i)),
  \end{equation}
  \end{small}
  where $U(a_i)$ is a comprehensive score for quantifying the DM's perceived attractiveness of alternative $a_i$, and $u_j(g_j(a_i))$ is a marginal value of alternative $a_i$ on each criterion $g_j$, $j=1,\ldots,m$.
  Marginal value functions $u_j(\cdot)$ are supposed to be monotone (non-decreasing for gain-type criteria) and normalized such that the comprehensive value $U(a_i)$ is bound within the interval [0,1].
  
  The true value function model $U(\cdot)$ of the DM remains unknown a priori.
  Hence, we need a technique to approximate the true value function model $U(\cdot)$.
  There are diverse types of marginal value functions (including linear, piecewise-linear, splined, and general monotone ones) that can be utilized for this approximation (see \citep{liu2020preference}).
  For simplicity, we only consider the case of piecewise-linear marginal value functions.
  Approximating the true value function model using piecewise-linear marginal value functions can be seen as a nonparametric technique because we do not need to make any assumptions on the form of the true marginal value functions.
  Such an approximation technique has been widely used in the MCDA field (see, e.g., \citep{liu2023contingent}) and we present the detailed elaboration on this technique in Online Appendix D.
  With the approximation using piecewise-linear marginal value functions, the true value function model $U(\cdot)$ can be specified as
  \begin{small}
  \begin{equation}
  U\left(a_i\right)=\bm{u}^\mathrm{T} \bm{V}\left(a_i\right),
  \end{equation} 
  \end{small}
  
  where $\bm{u}$ is a $\gamma$-dimensional parameter vector to signify the intrinsic character of the preference model $U(\cdot)$, and $\bm{V}\left(a_i\right)$ is a $\gamma$-dimensional characteristic vector that is fully determined by the performances of alternative $a_i$ on multiple criteria, particularly irrelevant to the DM's preferences.
  In other words, the DM's value function model $U(\cdot)$ is parameterized by the preference vector $\bm{u}$.
  
  The following linear constraints are imposed to ensure the monotonicity of the marginal value functions $u_{j}(\cdot)$ and normalize the comprehensive value $U(\cdot)$ within the interval [0,1] (see Online Appendix D for more details):
  \begin{small}
  \begin{equation}
  \left\{\begin{array}{l}
  \bm{u} \geqslant \bm{0}, \\
  \bm{1} \cdot \bm{u}= 1,
  \end{array}\right.
  \end{equation}
  \end{small}
  where $\bm{1}$ and $\bm{0}$ are the vectors with all entries being equal to 1 and 0, respectively.
  One can observe that $\bm{u}$ resides in the $(\gamma-1)$-dimensional simplex.
  In light of such constraints, it is reasonable to specify both the prior of the preference model $p(\bm{u} \mid \bm{\alpha})$ and the variational distribution $q(\bm{u} \mid \bm{\theta})$ as two Dirichlet distributions:
  \begin{small}
	  \begin{equation}
			  p(\bm{u} \mid \bm{\alpha})=\mathrm{Dir}(\bm{u} \mid \bm{\alpha})\quad \mbox{and}\quad q(\bm{u} \mid \bm{\theta})=\mathrm{Dir}(\bm{u} \mid \bm{\theta}),
	  \end{equation}
	  \end{small}
  where 
  $\bm{\alpha}=(\alpha_1, \dots, \alpha_\gamma)^\mathrm{T} \in \mathbb{R}^\gamma$ and $\bm{\theta}=(\theta_1, \dots, \theta_\gamma)^\mathrm{T} \in \mathbb{R}^\gamma$ are the parameters for the two Dirichlet distributions such that $\alpha_k > 0$ and $\theta_k > 0$ for $k=1, \dots, \gamma$.
  Note that the Dirichlet distributions are defined over the $(\gamma-1)$-dimensional simplex, and thus, the non-negativity and normalization of $\bm{u}$ can be fulfilled.
  In particular, we set $\bm{\alpha} = \bm{1}$ in this paper so that the prior $p(\bm{u} \mid \bm{\alpha})$ boils down to the uniform distribution over the $(\gamma-1)$-dimensional simplex, which implies that we have no other reasons to favor one preference model over another since no preference information from the DM is observed before starting the interactive preference elicitation procedure.
  
  Once the prior $p(\bm{u} \mid \bm{\alpha})$ and the variational distribution $q(\bm{u} \mid \bm{\theta})$ are specified, we can apply the proposed Bayesian interactive preference elicitation framework to interact with the DM to progressively capture the DM's preferences in the MCDA context.
  As a key component of the proposed framework, the approximation inference approach proposed in Section \ref{sec:Variational Bayesian Approach} will be repeatedly invoked to use a variational distribution $q(\bm{u} \mid \bm{\theta})$ to approximate the posterior $p(\bm{u} \mid Q^{(t)})$ for $t=1,\ldots,T$.
  The effectiveness and efficiency of the approximation inference approach influence the outcome of capturing the DM's preferences to a large extent.
  In the following, let us investigate more implementation details of the approximation inference approach in the MCDA context.
  
  Recall that the approximation inference approach maximizes the ELBO function $L\left(\bm{\theta}\right)$ to find the optimal variational distribution $q(\bm{u} \mid \bm{\theta}^*(Q^{(t)}))$ that can minimize the KL divergence $D_{\mathbb{KL}}\left(q\left(\bm{u} \mid \bm{\theta}\right) \mid\mid p\left(\bm{u} \mid Q^{(t)}\right)\right)$.
  We use a gradient-based optimization algorithm to address the problem based on the gradient formulated in Equation (\ref{eq:5}).
  In the MCDA context, the variational distribution $q(\bm{u} \mid \bm{\theta})$ is a Dirichlet distribution $\mathrm{Dir}(\bm{u} \mid \bm{\theta})$, which requires $\bm{\theta} > \bm{0}$.
  To satisfy this requirement in the gradient-based optimization algorithm, we can introduce an auxiliary vector $\bm{\phi} = \left(\phi_1,\ldots,\phi_\gamma\right)^\mathrm{T} \in \mathbb{R}^\gamma$ such that $\theta_k = \phi_k^2$ for $k=1,\ldots,\gamma$.
  We denote this relationship as $\bm{\theta} = \bm{\phi}^2$.
  In this way, the gradient in Equation (\ref{eq:5}) can be expressed in terms of $\bm{\phi}$ as:
  
  \begin{small}
  \begin{equation}\label{eq:17}
		   \nabla_{\bm{\phi}}L(\bm{\phi}) \approx \sum_{w=1}^W [\nabla_{\bm{\phi}^2}\log q(\bm{u}^{(w)} \mid \bm{\phi}^2)][\log p(Q^{(t)} \mid \bm{u}^{(w)}) + \log p(\bm{u}^{(w)} \mid \bm{\alpha})-\log q(\bm{u}^{(w)} \mid \bm{\phi}^2)].
  \end{equation}
  \end{small}
  Note that the gradient $\nabla_{\bm{\phi}}L(\bm{\phi})$ is approximated using $W$ samples of the parameter $\bm{u}$ drawn from the distribution $q(\bm{u} \mid \bm{\phi}^2)$.
  A practical concern for this approximation is that its variance may be very large due to the existence of the term $\nabla_{\bm{\phi}^2}\log q(\bm{u} \mid \bm{\phi}^2)$ (\cite{paisley2012variational}). 
  As a result, a considerable number of samples are necessary to reduce this variance to a level suitable for approximating the gradient, implying an increased computational requirement.
  
  To mitigate the issue of high variance, we opt for a useful method known as the reparameterization trick (RT) (\cite{kingma2013auto}) for sampling from $q(\bm{u}\mid \bm{\phi}^2)$.  
  The essence of RT is to compute a sample from $q(\bm{u}\mid \bm{\phi}^2)$ by first sampling an auxiliary variable $\bm{\epsilon}$ from some noise distribution $p(\bm{\epsilon})$ which is independent of $\bm{\phi}$, and then transforming $\bm{\epsilon}$ to $\bm{u}$ using a deterministic and differentiable vector-valued function $g_{\bm{\phi}}(\cdot)$ parameterized by $\bm{\phi}$ (\cite{murphy2023probabilistic}).
  This way, we can derive a lower variance estimator for the gradient in Equation (\ref{eq:17}).
  Such a reparameterization trick has been widely used in variational inference-related studies (see \cite{mohamed2020monte} for a review) and deep generative models (e.g., Variational AutoEncoder (VAE) (\cite{kingma2013auto})).
  
  In the MCDA context, the auxiliary variable $\bm{\epsilon}$ can be a Gaussian noise such that $\bm{\epsilon} \sim \mathcal{N}(\bm{0}, \bm{I})$, where $\bm{I}$ is $\gamma$-dimensional identity matrix.
  According to \cite{joo2020dirichlet}, the reparameterization trick to transform from $\bm{\epsilon}$ to $\bm{u}$ is given as
  \begin{small}
  \begin{equation}
  \bm{u} = g_{\bm{\phi}}(\bm\epsilon)=\mathrm{softmax}(\bm\mu + \sqrt{\bm\Sigma} \times \bm\epsilon),
  \end{equation}
  \end{small}
  where $\bm{\mu} = (\mu_1, \ldots, \mu_\gamma)^\mathrm{T} \in \mathbb{R}^\gamma$ with each entry $\mu_k = \log{\phi^2_k}-\frac{1}{\gamma}\sum_{i=1}^{\gamma} \log{\phi^2_i}$, and $\bm\Sigma \in \mathbb{R}^{\gamma \times \gamma}$ is a diagonal matrix with each entry in the diagonal $\Sigma_k = \frac{1}{\phi^2_k}\left(1 - \frac{2}{\gamma}\right) + \frac{1}{{\gamma}^2}\sum_{i=1}^{\gamma} \frac{1}{\phi^2_i}$, for $k=1,\ldots,\gamma$.
  In this way, $\bm{u}$ follows a Dirichlet distribution $q(\bm{u}\mid \bm{\phi}^2)$.
  
  On the basis of RT, the gradient in Equation (\ref{eq:17}) can be reformulated as:
  \begin{small}
  \begin{equation} \label{eq:19}
	   \begin{aligned}
				 \nabla_{\bm\phi}L(\bm\phi)
				 &=\nabla_{\bm\phi} \int q(\bm{u}\mid \bm\phi^2)\log \frac{p(Q^{(t)}, \bm u \mid \bm{\alpha})}{q(\bm{u} \mid \bm\phi^2)} \mathrm{d}\bm{u}\\
				 &=\nabla_{\bm\phi} \int [\log p(Q^{(t)}, \bm{u} \mid \bm\alpha)-\log q(\bm{u} \mid \bm\phi^2)] q(\bm{u} \mid \bm\phi^2)\mathrm{d}\bm{u}\\
				 &=\nabla_{\bm\phi} \int [\log p(Q^{(t)}, \bm{u} \mid \bm\alpha)-\log q(\bm{u} \mid \bm\phi^2)] p(\bm\epsilon) \mathrm{d}\bm\epsilon\\
				 &=\nabla_{\bm\phi} \mathbb{E}_{\bm{\epsilon} \sim p(\bm\epsilon)}\left[\log p(Q^{(t)}, \bm{u} \mid \bm\alpha)-\log q(\bm{u} \mid \bm\phi^2)\right]\\
				 &= \mathbb{E}_{\bm{\epsilon} \sim p(\bm\epsilon)}\left[\nabla_{\bm\phi} \left(\log p(Q^{(t)}, \bm{u} \mid \bm\alpha)-\log q(\bm{u} \mid \bm\phi^2)\right) \right].
	   \end{aligned}
  \end{equation}
  \end{small}
  Subsequently, by incorporating $\bm{u} = g_{\bm{\phi}}(\bm\epsilon)=\mathrm{softmax}(\bm\mu + \bm\Sigma \times \bm\epsilon)$, we further derive:
  \begin{small}
  \begin{equation} \label{eq:20}
	  \begin{aligned}
				\nabla_{\bm\phi}L(\bm\phi)
				&= \mathbb{E}_{\bm{\epsilon} \sim p(\bm\epsilon)}\left[\nabla_{\bm\phi} \left(\log p(Q^{(t)}, \bm{u} \mid \bm\alpha)-\log q(\bm{u} \mid \bm\phi^2)\right) \right]\\
				&=\mathbb{E}_{\bm{\epsilon} \sim p(\bm\epsilon)}\left[\nabla_{\bm{u}} \left(\log p(Q^{(t)}, \bm{u} \mid \bm\alpha)-\log q(\bm{u} \mid \bm\phi^2)\right) \nabla_{\bm\phi}\bm{u} \right]\\
				&= \mathbb{E}_{\bm{\epsilon} \sim p(\bm\epsilon)}\left[\nabla_{\bm{u}} \left(\log p(Q^{(t)}, \bm{u} \mid \bm\alpha)-\log q(\bm{u} \mid \bm\phi^2)\right) \nabla_{\bm\phi}\left(\mathrm{softmax}(\bm\mu + \bm\Sigma \times \bm\epsilon)\right) \right]\\
				&\approx \frac{1}{W} \sum_{w=1}^{W}\left[\nabla_{\bm{u}} \left(\log p(Q^{(t)}, \bm{u} \mid \bm\alpha)-\log q(\bm{u} \mid \bm\phi^2)\right) \nabla_{\bm\phi}\left(\mathrm{softmax}(\bm\mu + \bm\Sigma \times \bm\epsilon^{(w)})\right) \right],
	  \end{aligned}
  \end{equation}
  \end{small}
  \noindent where $\bm{\epsilon}^{(w)}$, $w=1,\ldots,W$, are samples of the auxiliary variable $\bm{\epsilon}$ drawn from the Gaussian distribution $\mathcal{N}(\bm{0}, \bm{I})$ and $\log p(Q^{(t)}, \bm{u} \mid \bm\alpha)=\log p(Q^{(t)} \mid \bm{u}) + \log p(\bm{u} \mid \bm{\alpha})$. 
  Next, with the lower variance estimator for the gradient in Equation (\ref{eq:20}), we can use a gradient-based optimization algorithm to find the optimal variational distribution $q(\bm{u} \mid \bm{\theta}^*(Q^{(t)}))$ for approximating the posterior $p(\bm{u} \mid Q^{(t)})$.
  
  In the following two sections, we validate the performance of the proposed interactive preference elicitation framework in MCDA ranking problems. The experimental study is partitioned into two parts. The first is to assess the efficacy of the proposed variational Bayesian approach in reproducing adequate relations for all pairs of alternatives. The other is to explore the performance of the MCTS-based questioning policy in terms of reducing the variance of the posterior distribution and the uncertainty regarding the recommended ranking. The experimental study is carried out on a server (Intel Xeon Silver 4214 CPU 2.20GHz, RAM 128G, SSD 480G, 4$\times$RTX 2080Ti 11G), employing Python 3.8.

  \section{Computational Experiments on Preference Inference}\label{sec:Experiments on Preference Inference}
  In this section, we evaluate the performance of the proposed variational Bayesian approach on real-world datasets across diverse MCDA problem settings, focusing on the accuracy of preference model inference.
  
  We begin by outlining the experimental design, which includes the baseline method, Stochastic Ordinal Regression (SOR), and the evaluation metric, \textit{Average Support of the Inferred Pairwise Outranking Indices (ASP)}, used to assess the performance of preference estimation methods. Furthermore, we introduce the real-world datasets employed in this study, along with the problem settings characterized by three key factors: the shape of the marginal value function ($F_1$), the number of pairwise comparisons provided by the decision-maker ($F_2$), and the proportion of biased preference information ($F_3$). Detailed descriptions are provided in Online Appendix E.
  
  We then present and analyze the experimental results in detail. The average ASP values for the proposed variational Bayesian inference approach (both with and without RT) and SOR, across different levels of factors ($F_1$, $F_2$, $F_3$) and real-world datasets, are reported in Table \ref{tab:ASP_results}. The RT-enhanced variant consistently achieves higher ASP values across all datasets, demonstrating superior performance over the baseline methods. The Wilcoxon test (see Online Appendix F) confirms the statistical significance of these results, with $p$-values below 0.001 for most datasets, further validating the advantage of our RT-based approach.
  \begin{table}[t]
	  \centering\footnotesize
	  \caption{Average ASP for the variants of the proposed variational Bayesian inference approach with/without RT and SOR derived from the experiment (Bolded numbers indicate the best result among the three methods.).}
	  \setlength{\abovecaptionskip}{0.05cm}
	  \centering
	  \renewcommand\arraystretch{0.7}
	  \label{tab:ASP_results}
	  \resizebox{1\linewidth}{!}{
		  \begin{tabular}{*{15}{c}}
			  \toprule
			  \multirow{2}{*}{Dataset} & \multirow{2}{*}{Method}               & \multicolumn{4}{c}{F1} & \multicolumn{4}{c}{F2} & \multicolumn{4}{c}{F3} & \multirow{2}{*}{AVG}                                                                                                                                                          \\
			  \cmidrule(lr){3-6} \cmidrule(lr){7-10} \cmidrule(lr){11-14}
									   &                                       & linear                 & concave                & convex                 & mixture              & 20             & 40             & 60             & 80             & 0              & 10\%           & 20\%           & 30\%                            \\
			  \midrule
			  \multirow{3}{*}{TC}      & \makecell{The variant with RT}
									   & \textbf{0.901}                        & \textbf{0.900}         & \textbf{0.884}         & \textbf{0.897}         & \textbf{0.874}       & \textbf{0.895} & \textbf{0.904} & \textbf{0.909} & 0.936          & \textbf{0.907} & \textbf{0.882} & \textbf{0.856} & \textbf{0.895}                  \\
									   & \makecell{The counterpart without RT}
									   & 0.895                                 & 0.897                  & 0.876                  & 0.888                  & 0.868                & 0.888          & 0.898          & 0.903          & 0.924          & 0.900          & 0.879          & 0.853          & 0.889                           \\
									   & SOR
									   & 0.883                                 & 0.881                  & 0.873                  & 0.879                  & 0.864                & 0.882          & 0.887          & 0.883          & \textbf{0.959} & 0.906          & 0.853          & 0.799          & 0.879                           \\
			  \multirow{3}{*}{IC}      & \makecell{The variant with RT}        & 0.962                  & \textbf{0.957}         & \textbf{0.961}         & \textbf{0.960}       & 0.953          & \textbf{0.959} & \textbf{0.963} & \textbf{0.965} & 0.970          & \textbf{0.964} & 0.956          & 0.950          & \textbf{0.960} \\
									   & \makecell{The counterpart without RT} & \textbf{0.963}         & 0.956                  & 0.960                  & 0.958                & \textbf{0.954} & 0.958          & 0.962          & 0.963          & 0.967          & 0.962          & \textbf{0.957} & \textbf{0.952} & 0.959          \\
									   & SOR                                   & 0.957                  & 0.950                  & 0.952                  & 0.951                & 0.950          & 0.952          & 0.954          & 0.953          & \textbf{0.974} & 0.960          & 0.945          & 0.930          & 0.952          \\
			  \multirow{3}{*}{SS}      & \makecell{The variant with RT}        & \textbf{0.943}         & \textbf{0.935}         & \textbf{0.937}         & \textbf{0.940}       & \textbf{0.927} & \textbf{0.938} & \textbf{0.944} & \textbf{0.947} & 0.947          & \textbf{0.944} & \textbf{0.937} & \textbf{0.928} & \textbf{0.939} \\
									   & \makecell{The counterpart without RT} & 0.941                  & 0.931                  & 0.935                  & 0.937                & 0.926          & 0.935          & 0.940          & 0.943          & 0.943          & 0.941          & 0.935          & 0.927          & 0.936          \\
									   & SOR                                   & 0.937                  & 0.927                  & 0.930                  & 0.932                & 0.924          & 0.931          & 0.934          & 0.937          & \textbf{0.949} & 0.941          & 0.927          & 0.910          & 0.932          \\
			  \multirow{3}{*}{CE}      & \makecell{The variant with RT}        & \textbf{0.899}         & \textbf{0.886}         & \textbf{0.887}         & \textbf{0.890}       & \textbf{0.865} & \textbf{0.891} & \textbf{0.901} & \textbf{0.907} & 0.904          & \textbf{0.898} & \textbf{0.889} & \textbf{0.872} & \textbf{0.891} \\
									   & \makecell{The counterpart without RT} & 0.894                  & 0.879                  & 0.877                  & 0.880                & 0.858          & 0.882          & 0.892          & 0.898          & 0.894          & 0.889          & 0.881          & 0.867          & 0.883          \\
									   & SOR                                   & 0.891                  & 0.875                  & 0.883                  & 0.883                & 0.856          & 0.883          & 0.893          & 0.900          & \textbf{0.906} & 0.897          & 0.878          & 0.852          & 0.883          \\
			  \multirow{3}{*}{RU}      & \makecell{The variant with RT}        & \textbf{0.942}         & \textbf{0.935}         & \textbf{0.936}         & \textbf{0.936}       & \textbf{0.924} & \textbf{0.935} & \textbf{0.943} & \textbf{0.947} & 0.944          & \textbf{0.943} & \textbf{0.936} & \textbf{0.926} & \textbf{0.937} \\
									   & \makecell{The counterpart without RT} & 0.940                  & 0.929                  & 0.932                  & 0.930                & 0.921          & 0.931          & 0.937          & 0.941          & 0.939          & 0.937          & 0.932          & 0.923          & 0.933          \\
									   & SOR                                   & 0.935                  & 0.930                  & 0.923                  & 0.929                & 0.918          & 0.926          & 0.934          & 0.939          & \textbf{0.948} & 0.939          & 0.924          & 0.907          & 0.929          \\
			  \multirow{3}{*}{EBC}     & \makecell{The variant with RT}        & \textbf{0.886}         & \textbf{0.870}         & \textbf{0.871}         & \textbf{0.875}       & \textbf{0.842} & \textbf{0.874} & \textbf{0.889} & \textbf{0.898} & 0.891          & \textbf{0.884} & \textbf{0.873} & \textbf{0.855} & \textbf{0.875} \\
									   & \makecell{The counterpart without RT} & 0.881                  & 0.863                  & 0.864                  & 0.868                & 0.838          & 0.867          & 0.881          & 0.890          & 0.881          & 0.875          & 0.868          & 0.852          & 0.869          \\
									   & SOR                                   & 0.882                  & 0.867                  & 0.861                  & 0.867                & 0.836          & 0.866          & 0.882          & 0.893          & \textbf{0.892} & 0.882          & 0.866          & 0.837          & 0.869          \\
			  \multirow{3}{*}{MPI}     & \makecell{The variant with RT}        & \textbf{0.894}         & \textbf{0.875}         & \textbf{0.887}         & \textbf{0.882}       & \textbf{0.860} & \textbf{0.881} & \textbf{0.894} & \textbf{0.903} & 0.901          & \textbf{0.894} & \textbf{0.880} & \textbf{0.863} & \textbf{0.885} \\
									   & \makecell{The counterpart without RT} & 0.892                  & 0.867                  & 0.882                  & 0.877                & 0.858          & 0.876          & 0.888          & 0.896          & 0.893          & 0.887          & 0.876          & 0.862          & 0.879          \\
									   & SOR                                   & 0.888                  & 0.867                  & 0.878                  & 0.872                & 0.854          & 0.875          & 0.883          & 0.893          & \textbf{0.907} & 0.893          & 0.867          & 0.838          & 0.876          \\
			  \multirow{3}{*}{GUU}     & \makecell{The variant with RT}        & \textbf{0.884}         & \textbf{0.867}         & \textbf{0.866}         & \textbf{0.866}       & \textbf{0.841} & \textbf{0.871} & \textbf{0.882} & \textbf{0.889} & \textbf{0.888} & \textbf{0.878} & \textbf{0.868} & \textbf{0.850} & \textbf{0.871} \\
									   & \makecell{The counterpart without RT} & 0.882                  & 0.858                  & 0.858                  & 0.861                & 0.840          & 0.864          & 0.874          & 0.881          & 0.878          & 0.870          & 0.863          & 0.849          & 0.865          \\
									   & SOR                                   & 0.879                  & 0.858                  & 0.858                  & 0.858                & 0.839          & 0.864          & 0.872          & 0.879          & 0.887          & 0.875          & 0.858          & 0.833          & 0.863          \\
			  \bottomrule
		  \end{tabular}
	  }
  \end{table}
  The following subsections provide an in-depth analysis of performance variations across different datasets and experimental settings.
  
  \subsection{Performance analysis for different shapes of marginal value functions} \label{sec:Performance_F1}
  
\begin{figure}[htbp]
	  \centering
	  \begin{subfigure}[b]{0.32\textwidth}
		  \centering
		  \includegraphics[width=\textwidth]{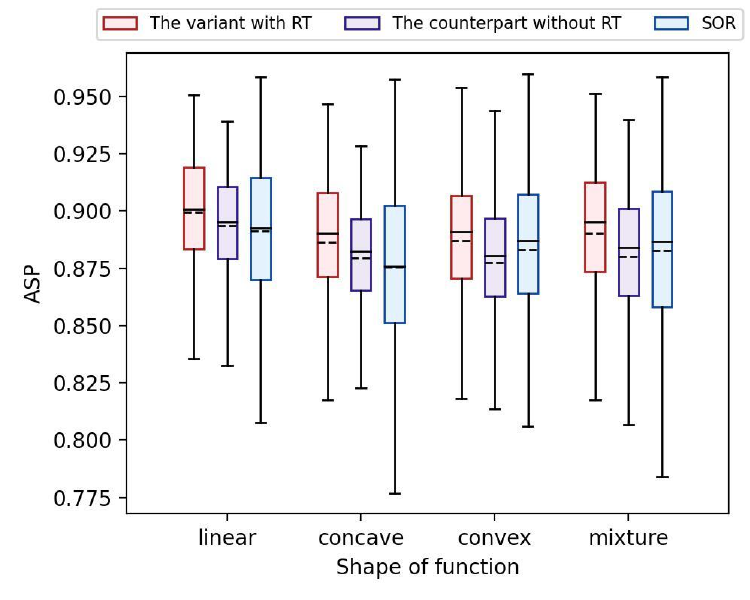}
		  \caption{}
		  \label{fig:subfig1}
	  \end{subfigure}
	  \hfill
	  \begin{subfigure}[b]{0.32\textwidth}
		  \centering
		  \includegraphics[width=\textwidth]{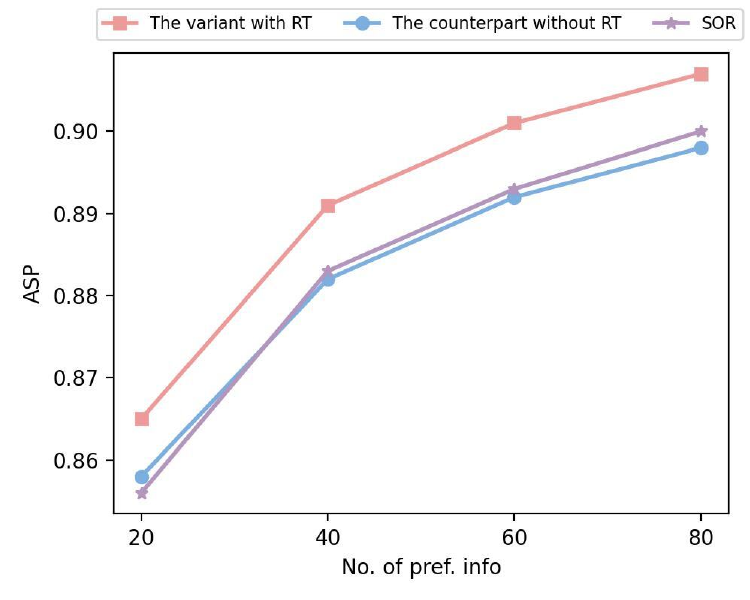}
		  \caption{}
		  \label{fig:subfig2}
	  \end{subfigure}
	  \hfill
	  \begin{subfigure}[b]{0.32\textwidth}
		  \centering
		  \includegraphics[width=\textwidth]{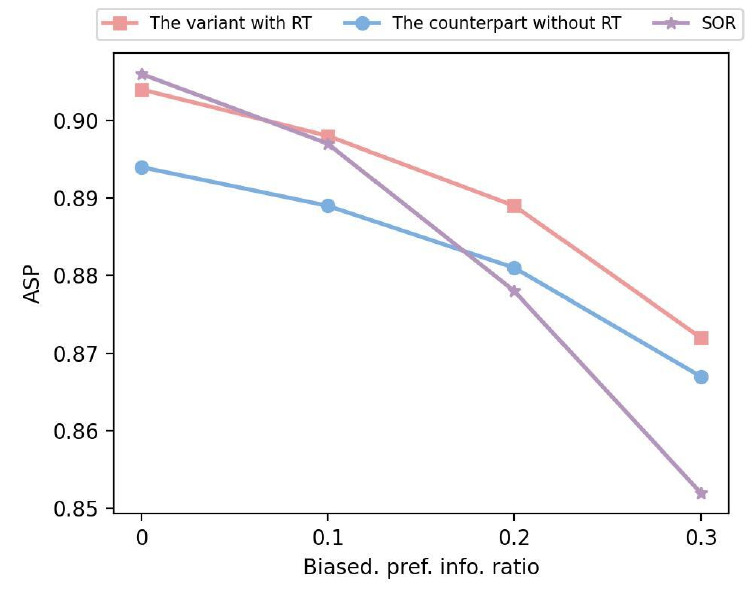}
		  \caption{}
		  \label{fig:subfig3}
	  \end{subfigure}
	  \vspace{-5pt} 
	  \caption{Average ASP for the variants of the proposed variational Bayesian inference approach with/without RT and SOR, evaluated on the CE dataset, considering various shapes of marginal value functions, different numbers of pairwise comparisons, and varying proportions of biased preference information.}
	  \label{fig:three_subfigures_subcaption}
  \end{figure}

  This section analyzes the experimental results across different datasets, focusing on various shapes of marginal value functions. Due to space limitations, Figure \ref{fig:subfig1} presents boxplots of the average ASP values for a single dataset, CE, while the complete results for all datasets, which exhibit similar trends, are available in Online Appendix G. By examining the mean and median values, it is evident that the proposed method with RT consistently outperforms the other two methods regardless of the form of the marginal value function. This advantage holds across all four types of marginal value functions. Furthermore, the results for the two variants of the proposed variational Bayesian inference approach are more concentrated, indicating greater stability in inferring DMs' preferences. Finally, ASP values are higher for all three methods when the marginal value function is linear, which is expected since we approximate the true function using a piecewise linear form. Simpler functions require fewer sub-intervals and parameters, making them easier to estimate, whereas more complex functions demand more sub-intervals and parameters, increasing the estimation challenge.
  
  \vspace{-5pt}
  
  \subsection{Performance analysis for different numbers of pairwise comparisons} \label{sec:Performance_F2}
  
  This section examines the impact of varying amounts of preference information on inference performance. We compare the proposed variational Bayesian inference approach (with and without RT) and SOR under $\{20, 40, 60, 80\}$ pairwise comparisons. Figure \ref{fig:subfig2} presents results for the CE dataset, while the complete results for all datasets are provided in Online Appendix G. As the number of comparisons increases, all three methods exhibit gradual performance improvements, as additional preference information reduces uncertainty in inferring the DM's preference model. Notably, SOR's performance is more sensitive to the amount of preference information compared to our Bayesian approach. This is due to fundamental differences between the two methods: SOR encodes preference constraints as linear inequalities, refining the feasible region as more information is provided. In contrast, our Bayesian approach models preferences probabilistically, capturing variations across value functions. Consequently, even with limited data, our method infers a well-concentrated posterior distribution, reducing uncertainty more effectively. 
   
  Interestingly, the ASP values of the SOR method decrease when the number of pairwise comparisons increases from 60 to 80 for the first two datasets (TC and IC). This anomaly arises because a larger amount of preference information also introduces more biased inputs. Since SOR constructs feasible regions based on linear constraints, biased inputs may exclude the true preference model, compromising its performance. In contrast, our Bayesian approach assigns probabilities to all possible preference models, mitigating the risk of discarding the true model due to bias. Section \ref{sec:Performance_F3} provides more direct evidence for this point. Moreover, as the amount of accurate preference information increases, it helps to correct the impact of biased data, which is consistent with the conclusions drawn from the results.
  
  \subsection{Performance analysis for different proportions of biased preference information} \label{sec:Performance_F3}
  
  In this section, we examine the performance of the variants of the proposed variational Bayesian inference approach with/without RT and SOR with biased preference information. Figure \ref{fig:subfig3} presents results for the CE dataset, while the complete results for all datasets are provided in Online Appendix G. Empirical results can be summarized from the following two aspects. On the one hand, as the proportion of biased preference information increases, the performance of all three methods declines, which is in line with our expectations. This is because an increase in biased preference information causes the posterior distribution of the value function to diverge from the true preference model $U^{true}(a)$ for the proposed Bayesian methods. As for SOR, it causes the feasible region of the preference model to be wrongly restricted. On the other hand, the magnitude of the decline in ASP with increasing biased preference information is less significant with our approach compared to SOR, indicating that the proposed variational Bayesian approach exhibits greater robustness to biased preference information. The reason is that our method assigns probabilities to all potential value function models, ensuring that the actual DM's preference model remains within the consideration set despite the presence of biased information. Moreover, by modeling the provided preferences from a probabilistic perspective, our approach ensures that even if a small proportion of comparisons is biased, the majority of consistent preference information will guide the posterior distribution of the value function closer to the assumed true preference model. 
  
  \subsection{Comparison of the variants of the proposed variational Bayesian inference approach with/without RT in high variance problem} \label{sec:Performance_RT}
  
  In this section, we compare the performance of the proposed variational Bayesian inference approach with and without RT and analyze the underlying causes of their differences. Our results show that while both variants exhibit similar trends across different problem settings ($F_1$, $F_2$, $F_3$), the variant with RT consistently outperforms the one without it. This aligns with expectations, as the latter suffers from high variance issues, leading to instability in gradient computation.
  
  To further examine RT's role in variance reduction, we conducted an additional experiment, computing gradients 100 times per gradient ascent iteration for both variants and calculating their variance across dimensions. Results, averaged over iterations, problem settings, and dimensions, confirm that the variant with RT yields significantly lower gradient variance than its counterpart. For instance, in the first dataset, the median gradient variance for the RT variant is 0.02, compared to 7.35 without RT. These findings underscore RT's effectiveness in stabilizing inference. For a more detailed presentation of results, including visualizations, see Online Appendix G.
  
  \subsection{Summary of experiments on preference inference} \label{sec:Summary of experiments on preference inference}
  The experimental analysis demonstrates the superiority of the proposed variational Bayesian inference approach with RT compared to the counterpart without RT and the SOR method. On the one hand, the variant of the proposed variational Bayesian inference approach with RT outperforms both the counterpart without RT and SOR in terms of ASP across different datasets and problem settings, including the diversity of the shape of the marginal value function, the number of pairwise comparisons, and the proportion of biased preference information. On the other hand, the experimental results show that the variant with RT effectively mitigates the high variance issue, leading to a more accurate inference of the DM's preferences. These features make our approach easy to implement in practice and increase its potential to produce credible decision outcomes.
  
  \section{Computational Experiments on Questioning Policy}\label{sec:Experiments on Questioning Policy}
  This section presents the results of applying the proposed questioning policy to the preference elicitation problem in the context of MCDA. We conduct a comprehensive evaluation to demonstrate the practical effectiveness of our approach in efficiently constructing the DM's preference model.
  
  We begin by outlining the experimental design, including the evaluation metrics—$f_\textit{VAR}$, $f_\textit{PWI}$, and $f_\textit{RAI}$—used to assess the performance of the questioning policy. Additionally, we introduce the baseline methods employed for comparison: $H_\textit{PWI}$, $H_\textit{RAI}$, $H_{\textit{PWI}-2}$, $H_{\textit{RAI}-2}$, $H_\textit{DVF}$, and $H_\textit{RAND}$. We also describe the problem settings, considering variations in the number of alternatives and criteria. Due to space limitations, detailed descriptions are provided in Online Appendix H.
  
\begin{figure}[!htbp]
	  \centering
	  \includegraphics[scale=0.35]{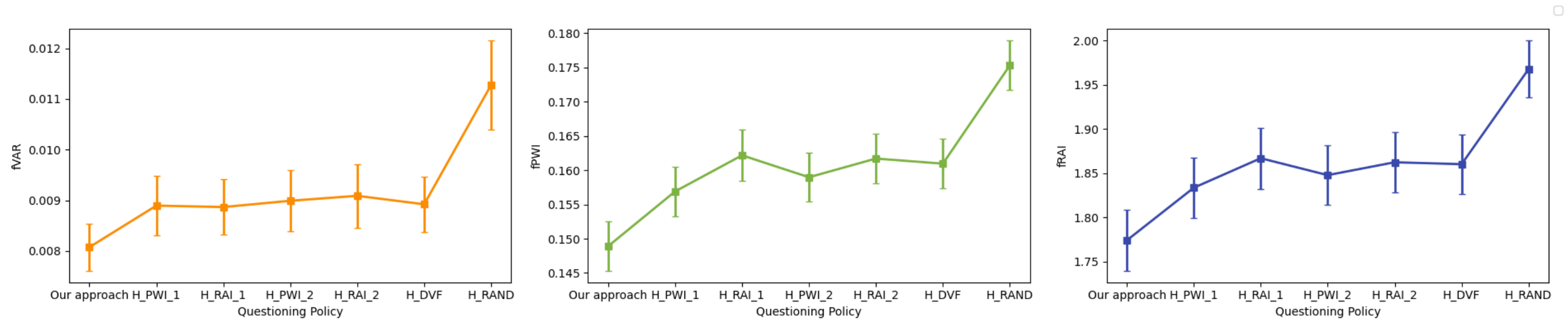}
	  \caption{Average values of different metrics characterizing the uncertainty of the DM's preferences with 7 questioning policies.} \label{fig:Avg_7policies}
  \end{figure}

  Next, we present the experimental results, comparing our MCTS-based questioning policy with the six baseline approaches. Figure \ref{fig:Avg_7policies} presents the average values of the evaluation metrics $f_\textit{VAR}$, $f_\textit{PWI}$, and $f_\textit{RAI}$ across all tested scenarios. As reported, our MCTS-based approach consistently achieves the lowest values for all three metrics, demonstrating its effectiveness in reducing uncertainty about the DM's preferences through interactions. The results of the one-sided Wilcoxon test (available in Online Appendix I) further confirm the statistical superiority of our method. In contrast, the worst-performing baseline, $H_\textit{RAND}$, exhibits significantly higher metric values, underscoring the necessity of active exploration strategies in preference elicitation.
  
  The following subsections provide a detailed analysis of the performance of these policies under varying numbers of alternatives and criteria, as well as an investigation into the impact of interaction rounds on the results.
  
  \subsection{Performance analysis for different numbers of alternatives and criteria} \label{sec:Performance analysis for different numbers of alternatives and criteria}
  
  The results in Online Appendix J presenting the metrics $f_\textit{VAR}$, $f_\textit{PWI}$, and $f_\textit{RAI}$, which characterize the uncertainty in the DM's preferences after $s$ interactions guided by different questioning policies, show consistent trends across problem settings. Specifically, for problems involving $n$ alternatives and $3$ criteria, and problems with $8$ alternatives and $m$ criteria, where $s \in {2, 4, 6, 8}$, $n \in {6, 7, 8, 9, 10}$, and $m \in {2, 3, 4, 5}$, our MCTS-based policy consistently outperforms other methods. In contrast, the random strategy consistently underperforms.
  
  The performances of these seven methods have the same trend with the variation of the problem
  settings. As the number of alternatives increases, $f_\textit{RAI}$ rises across all methods, reflecting greater uncertainty in ranking as more alternatives are added. However, $f_\textit{VAR}$ and $f_\textit{PWI}$ do not show a clear pattern in response to changes in the number of alternatives, as they measure posterior distribution variance and preference relation uncertainty, respectively, which are less sensitive to the number of alternatives. On the other hand, as the number of criteria increases, the performance of the seven questioning policies in terms of the three metrics does not exhibit a clear trend.
  
  \subsection{Performance analysis for different numbers of interaction rounds} \label{sec:Performance analysis for different numbers of interaction rounds}
  
  The average values of $f_\textit{VAR}$, $f_\textit{PWI}$, and $f_\textit{RAI}$ after 2, 4, 6, and 8 interactions with seven questioning policies illustrate how the number of interaction rounds affects the results. The observation is two-fold. On the one hand, as the number of interaction rounds increases, the results for all methods show a decline across all metrics. This aligns with the fact that more preference information reduces our uncertainty about the DM's preferences. On the other hand, with more interactions, the advantages of our method become more pronounced in most problem settings. This is because our method employs a decision-time planning approach, which allocates the limited computational budget to rollouts that simulate sequences starting from the current state, allowing for careful consideration of the long-term impacts of decisions while still focusing on the current state. As a result, it mitigates the shortsightedness issue found in comparative methods, with this advantage becoming more evident as the number of interaction rounds increases. Detailed results are provided in Online Appendix J.
  
  \subsection{Summary of experiments on questioning policy} \label{Summary of experiments on questioning policy}
  
  Based on the above results, it is evident that the proposed questioning policy consistently outperforms the other comparative methods across most problem settings and evaluation metrics. This suggests that our questioning policy is more effective at reducing the uncertainty of the DM's preferences within a limited number of interactions. Moreover, as the number of interactions increases, the advantages of our method become more pronounced. This demonstrates that our approach is well suited for considering the long-term impacts of current decisions, thereby alleviating the shortsightedness issue in other questioning strategies.
  
  \section{Conclusions}\label{sec:Conclusion}
  This paper focused on interactive preference construction problems, a critical domain where preferences are incrementally elicited through structured interactions, enabling the construction of models that accurately represent user priorities. We introduced a novel framework for interactive elicitation of holistic pairwise comparisons, addressing key challenges in inferring preferences and designing efficient, adaptive questioning policies across diverse domains. Our essential methodological contribution is three-fold.
  
  First, we developed a variational Bayesian approach to infer real-time participants' preferences. By framing preference inference within a Bayesian learning paradigm, we derived the posterior distribution of the preference model instead of opting to use a single deterministic model or assuming an arbitrary uniform distribution. This way, we effectively managed uncertainty from limited preference information and reduced cognitive biases, which is particularly valuable in the early stages of elicitation. Such an approach assures that the true preference model remains within consideration, even in the presence of inconsistencies. The use of variational inference enables rapid preference model updates, ensuring the process remains adaptive even to scenarios with extensive interactions and large state-action spaces, and is computationally efficient, particularly compared to traditional sampling-based methods.
  
  Second, we proposed an adaptive questioning policy that seeks to maximize the cumulative reduction in uncertainty over multiple interaction rounds. By modeling the process as a finite Markov decision problem and employing Monte Carlo Tree Search, the framework strategically selects questions to optimize information gain and significantly reduce uncertainty. To avoid exhaustively exploring the entire state-action space, it balances exploitation (selecting high-value questions based on current knowledge) and exploration (evaluating less certain options), mitigating the risk of short-sighted or myopic decisions. This aspect has been neglected in the vast majority of the existing questioning strategies. The novel definition of question value based on expected variance reduction directly aligns with the goal of refining the preference model, making the policy both intuitive and theoretically grounded. Moreover, integration with the efficient Bayesian inference method ensures that the search space is explored effectively, accelerating convergence toward the true preference model with minimal interaction rounds.
  
  Third, we applied the framework to the context of interactive elicitation of pairwise comparisons in Multiple Criteria Decision Aiding, emphasizing its practicality. Using an additive value function as the preference model, we addressed challenges of high variance in gradient estimation during variational Bayesian inference by incorporating the reparameterization trick from machine learning. This innovation enhances the robustness and computational efficiency of the inference process, enabling more accurate characterization of preferences in complex decision scenarios.
  
  The proposed framework was validated through extensive computational studies using real-world and synthetic datasets in MCDA contexts. Experimental results demonstrated the superiority of the variational Bayesian approach, particularly the variant with the reparameterization trick, in accurately capturing preferences. Specifically, the proposed adaptive questioning policy consistently surpassed baseline heuristics in reducing uncertainty within a limited number of interaction rounds. It also outperformed state-of-the-art methods, including the ones designed to mitigate shortsightedness, across various problem settings, highlighting its robustness in different MCDA contexts. The performance advantage of the MCTS-based policy was more prominent as the number of interaction rounds increased, which can be justified by its ability to consider long-term decision impacts. Collectively, these results underline the practical utility of the framework, offering a robust and efficient solution for interactive preference elicitation in MCDA.
  
  Building on the advances presented in this paper, several promising research directions emerge for further enhancing the field of interactive preference elicitation. First, extending the proposed framework to handle dynamic and evolving preferences could open new avenues. Preferences may change over time, and developing adaptive mechanisms to detect and integrate such changes would enhance the robustness of preference models. Second, incorporating multi-modal data sources, such as textual feedback or behavioral analytics, could complement traditional pairwise comparisons, enabling richer preference modeling. This would be particularly valuable in domains like healthcare or user-centric AI systems, where implicit and explicit signals often co-exist. Another vital direction is the development of scalable algorithms for high-dimensional settings, such as scenarios with large numbers of alternatives and criteria. While the current framework incorporates computationally efficient techniques, further optimization or hybrid methods could address even more computationally intensive contexts, such as real-time recommendation systems or group decision-making with multiple stakeholders providing potentially conflicting inputs. Moreover, integrating explainability into the questioning policy and preference inference process is increasingly important, especially in critical domains like finance or public policy, where DMs need to understand the rationale behind recommendations or elicitation strategies. Lastly, integrating reinforcement learning techniques beyond MCTS, such as deep reinforcement learning or transfer learning, could improve the adaptability of questioning policies. This would allow continuous learning across sessions or users, further refining the model's efficiency. Addressing these directions would advance the theoretical foundations of interactive preference elicitation and further enhance its applicability across diverse decision-making environments.
  
  \ACKNOWLEDGMENT{Jiapeng Liu acknowledges support from the National Natural Science Foundation of China (grant no. 72071155, 72471184). Mi{\l}osz Kadzi{\'n}ski was supported by the Polish National Science Center under the SONATA BIS project (grant no. DEC-2019/34/E/HS4/00045).}

  \bibliographystyle{apalike}
  \bibliography{mybibfile}

  \clearpage 

\setcounter{equation}{0}

\theAPPENDIXTITLE{Supplemental Material for "Preference Construction: A Bayesian Interactive Preference Elicitation Framework Based on Monte Carlo Tree Search"}
\rule{\textwidth}{1pt}
\begin{small}
	The supplementary material includes: a detailed presentation of the proposed Bayesian interactive preference elicitation framework (Appendix A), the derivation of the gradient of the ELBO function (Appendix B), an in-depth description of the implementation of the MCTS-based questioning policy (Appendix C), the process of approximating the value function model $U(\cdot)$ using piecewise-linear functions (Appendix D), a comprehensive explanation of the experimental design for the preference inference and questioning policy experiments (Appendices E and H), the results of the Wilcoxon test assessing the superiority of the RT-enhanced variational Bayesian preference inference method and the MCTS-based questioning policy over baseline methods (Appendices F and I), detailed computational results across different datasets for preference inference experiments (Appendices G), and visualized results of the computational experiments on the questioning policy (Appendices J). 
\end{small}
\rule{\textwidth}{1pt}

  \begin{APPENDICES}

	\section{Bayesian Interactive Preference Elicitation Framework} \label{app-A}
	In this section, we describe the proposed Bayesian interactive preference elicitation framework depicted in Figure \ref{fig:framework}. Its phases are as follows:

	\textit{Step 1: Data Input.} The procedure begins by defining the set of alternatives $A=\{a_1,\dots,a_n\}$ and specifying the maximum number of interaction rounds $T$. Set a round indicator $t=1$.

	\textit{Step 2: Preference inference.} If $t>1$, use the developed variational Bayesian approach to characterize the participant's preferences based on the preference information $Q^{(t-1)}$ accumulated up to $(t-1)$-th round and derive the posterior distribution $p(U \mid Q^{(t-1)})$ of the preference model $U(\cdot)$ (see Section 2.2). Otherwise, i.e., in the first round, let $Q^{(t-1)} = \emptyset$ and set the posterior distribution $p(U \mid Q^{(t-1)})$ be equal to the prior $p(U)$ (i.e., $p(U \mid Q^{(t-1)}) = p(U)$).

	\textit{Step 3: Checking stopping criterion.} Check whether $t = T$. If so, go to step 5. Otherwise, go to step 4.

	\textit{Step 4: Acquiring new pairwise comparison from the participant.} Apply the proposed MCTS-based questioning policy according to the derived posterior distribution $p(U \mid Q^{(t-1)})$ to select a new pair of alternatives $(a_i, a_j)$ (see Section 2.3). Present the new question $(a_i ? a_j)$ to the participant and ask him/her to make a pairwise comparison $q^{(t)}$ in the form of either $a_i \succ a_j$ or $a_j \succ a_i$. Collect the new pairwise comparison $q^{(t)}$ and update the set of accumulated preference information pieces by $Q^{(t)} = Q^{(t-1)} \cup \{q^{(t)}\}$. Let $t = t+1$ and go to step 2.

	\textit{Step 5: Results output.} Output the preference information gathered throughout the whole procedure, represented as $Q^{(T)}$, as well as the participant's preference model described by the posterior distribution $p(U \mid Q^{(T)})$.
	\begin{figure}[!htbp]
		\centering
		\includegraphics[scale=0.5]{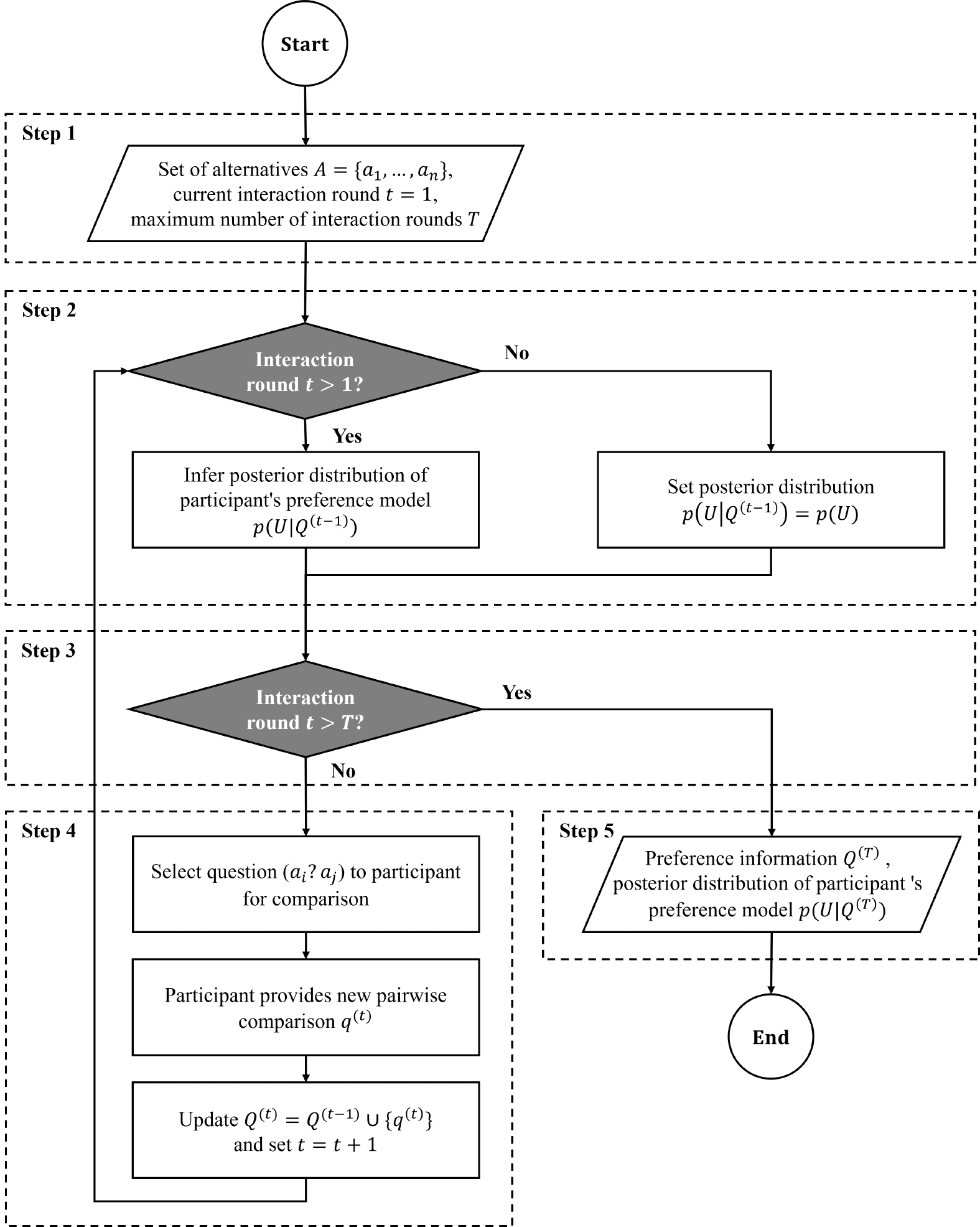}
		\caption{The proposed preference elicitation and model construction framework.} \label{fig:framework}
	\end{figure}
	
	\section{Derivation of the Gradient of the ELBO Function} \label{app-B}
	In this section, we detail the derivation of the gradient calculations outlined in Section 2.2, which enables the utilization of stochastic gradient ascent methodology to maximize the Evidence Lower Bound (ELBO) function, thereby yielding the optimized posterior distribution parameters $\bm\theta$. The gradient of the ELBO function $L(\bm\theta)$ with respect to the variational parameter vector $\bm{\theta}$ can be derived as follows: 
	\begin{footnotesize}
	\begin{equation}
		\begin{aligned}
			\nabla_{\bm{\theta}}L(\bm{\theta})
			& =\nabla_{\bm{\theta}} \int q(\bm{u}\mid \bm{\theta})\log \frac{p(Q^{(t)}, \bm u \mid \bm{\alpha})}{q(\bm{u} \mid \bm{\theta})} \mathrm{d}\bm{u}                                     \\
			& =\nabla_{\bm{\theta}} \int q(\bm{u} \mid \bm{\theta})[\log p(Q^{(t)}, \bm{u} \mid \bm\alpha)-\log q(\bm{u} \mid \bm{\theta})] \mathrm{d}\bm{u}                                      \\
			& =\int \nabla_{\bm{\theta}} \{ q(\bm{u} \mid \bm{\theta})[\log p(Q^{(t)}, \bm{u} \mid \bm\alpha)-\log q(\bm{u} \mid \bm{\theta})]\} \mathrm{d}\bm{u}                                 \\
			& =\underbrace {\int [\nabla_{\bm{\theta}} q(\bm{u} \mid \bm{\theta})][\log p(Q^{(t)}, \bm{u} \mid \bm\alpha)-\log q(\bm{u} \mid \bm{\theta})]\mathrm{d}\bm{u}}_{\mathrm{term\;A}} + \underbrace {\int q(\bm{u} \mid \bm{\theta})\nabla_{\bm{\theta}}[\log p(Q^{(t)}, \bm{u} \mid \alpha)-\log q(\bm{u} \mid \bm{\theta})]\mathrm{d}\bm{u}}_{\mathrm{term\;B}},
		\end{aligned}
	\end{equation}
	\end{footnotesize}
	where 
	\begin{footnotesize}
	\begin{equation}
		\begin{aligned}
			\text{term A}
			&=\int [\nabla_{\bm\theta} q(\bm{u} \mid \bm\theta)][\log p(Q^{(t)}, \bm{u} \mid \bm\alpha)-\log q(\bm{u} \mid \bm{\bm\theta})]\mathrm{d}\bm{u}\\
			&=\int [q(\bm{u} \mid \bm\theta) \nabla_{\bm\theta}\log q(\bm{u} \mid \bm\theta)][\log p(Q^{(t)}, \bm{u} \mid \bm\alpha)-\log q(\bm{u} \mid \bm\theta)]\mathrm{d}\bm{u}\\
			&=\int q(\bm{u} \mid \bm\theta) [\nabla_{\bm\theta}\log q(\bm{u} \mid \bm\theta)][\log p(Q^{(t)}, \bm{u} \mid \bm\alpha)-\log q(\bm{u} \mid \bm\theta)]\mathrm{d}\bm{u}\\
			&=\mathbb{E}_{q(\bm{u} \mid \bm\theta)}\{[\nabla_{\bm\theta}\log q(\bm{u} \mid \bm\theta)][\log p(Q^{(t)}, \bm{u} \mid \bm\alpha)-\log q(\bm{u} \mid \bm\theta)]\}\\
			&=\mathbb{E}_{q(\bm{u} \mid \bm\theta)}\{[\nabla_{\bm\theta}\log q(\bm{u} \mid \bm\theta)][\log p(Q^{(t)} \mid \bm{u}) + \log p(\bm{u} \mid \bm{\alpha})-\log q(\bm{u} \mid \bm\theta)]\},\\
		\end{aligned}
	\end{equation}
	\end{footnotesize}
	and
	\begin{footnotesize}
		\begin{equation}
			\begin{aligned}
				\text{term B}
				&=\int q(\bm{u} \mid \bm\theta)\nabla_{\bm\theta}[\log p(Q^{(t)}, \bm{u} \mid \bm\alpha)-\log q(\bm{u} \mid \bm\theta)]\mathrm{d}\bm{u}\\
				&=-\int q(\bm{u} \mid \bm\theta)\nabla_{\bm\theta}\log q(\bm{u} \mid \bm\theta)\mathrm{d}\bm{u}\\
				&=-\int \nabla_{\bm\theta} q(\bm{u} \mid \bm\theta)\mathrm{d}\bm{u}\\
				&=-\nabla_{\bm\theta} \int q(\bm{u} \mid \bm\theta)\mathrm{d}\bm{u}\\
				&=0.
				\end{aligned}
		\end{equation}
	\end{footnotesize}
	Thus, the gradient $\nabla_{\bm{\theta}}L(\bm{\theta})$ can be formulated as follows:
	\begin{equation*}
		\nabla_{\bm{\theta}}L(\bm{\theta})
		=\mathbb{E}_{\bm{u} \sim q(\bm{u} \mid \bm\theta)}\{[\nabla_{\bm\theta}\log q(\bm{u} \mid \bm\theta)][\log p(Q^{(t)} \mid \bm{u}) + \log p(\bm{u} \mid \bm{\alpha})-\log q(\bm{u} \mid \bm\theta)]\}.\\
		\end{equation*}

	\section{Monte Carlo Tree Search Algorithm} \label{app-C}
	This section provides a detailed description of the implementation of the MCTS-based questioning policy. 
	When the interactive preference elicitation process proceeds to each round $t$, the questioning policy $\pi \left(Q^{(t-1)}\right)$ implements an MCTS procedure to execute as many iterations as possible before a new question is posed to the participant, by incrementally building a question tree whose root node represents the current state and child nodes signify candidate questions that can be posed.
	Each iteration consists of the four operations:

	\begin{itemize}
		\item \textbf{Selection}. Starting in the root node, a child selection policy is recursively applied to traverse the tree, proceeding to the most urgent expandable node representing a particular question. 
		In our context, a node is expandable if both the following two conditions are satisfied: (a) it is a non-terminal node, i.e., the number of the involved questions from the root node to it does not reach $T-t+1$; (b) it has unexpanded children.
		
		\item \textbf{Expansion}. One potential question is added as a child node of the selected expandable node to expand the question tree.
		
		\item \textbf{Simulation}. A simulation is run from the new node to generate a complete sequence of $T$ questions and calculate the final variance reduction scale as a return after posing the $T$ questions.
		
		\item \textbf{Backpropagation}. The simulation result is backpropagated through the nodes along the generated question sequence to update their statistics.	
	\end{itemize}
	The pseudocode for the questioning policy \(\pi \left(Q^{(t-1)}\right)\) is presented in Algorithm \ref{alg:1}.
	The $\pi \left(Q^{(t-1)}\right)$ function iteratively invokes the SELECTION\&EXPANSION, SIMULATION, and BACKPROPAGATION functions to build a search tree until some predefined computational budget is reached.
	Computational budget is typically a time, memory, or iteration constraint (\cite{browne2012survey}), and in this paper, we specify it as a maximum number of iterations.
	In building the search tree, each node $node$ has two associated indicators: the total simulation reward $V(node)$ and the visit count $N(node)$.
	
	The SELECTION\&EXPANSION function combines the selection and expansion operations to find the most urgent expandable node in the following way: if node $node$ is non-terminal and not fully expanded, choose one of its child nodes to expand the tree; if node $node$ is non-terminal but fully expanded, recursively apply the BESTCHILD function to choose its best child until an expandable node is found.
	The BESTCHILD function determines how nodes in the tree are selected by treating the choice of the child node as a multi-armed bandit problem.
	Such a way effectively addresses the exploration-exploitation dilemma in MCTS by computing an Upper Confidence Bound (UCB) of child node $node^\prime$ as follows:

	\begin{small}
	\begin{equation} \label{eq:11}
		UCB(node^\prime) = \frac{V(node^\prime)}{N(node^\prime)} + 2 C_p \sqrt{\frac{2\ln N(node_0)}{N(node^\prime)}},
	\end{equation}
\end{small}

	where $N(node^\prime)$ and $N(node_0)$, respectively, count the number of times the child node $node^\prime$ and the root node $node_0$ have been visited, $V(node^\prime)$ is the the total simulation reward through $node^\prime$, and $C_p$ is a constant (usually set as ${1 \mathord{\left/
		{\vphantom {1 {\sqrt 2 }}} \right.
		\kern-\nulldelimiterspace} {\sqrt 2 }}$ (\cite{kocsis2006bandit})).
	For child node $node^\prime$, the first (exploitation) term is derived by averaging the accumulated rewards approximated by Monte Carlo simulations, whereas the second term (exploration) gives a measure of the uncertainty or variance in the estimate of its reward.
	Therefore, UCB balances exploitation of the currently most promising node with an exploration of others, which may later turn out to be superior (\cite{kocsis2006bandit}).
	The SIMULATION function applies Monte Carlo simulations to estimate the reward of the selected node by predicting the participant's preferences for unanswered questions in the constructed sequence of $T$ questions and calculating the variance reduction scale.
	For each candidate question $\left(a_i ? a_j\right)$, we predict the participant's preference in form of $a_i \succ a_j$ or $a_j \succ a_i$ according to the posterior predictive probabilities $p(a_i \succ a_j \mid Q^{(t-1)})$ and $p(a_j \succ a_i \mid Q^{(t-1)})$.
	The BACKPROPAGATION function updates the statistics for the nodes along the trajectory, starting from the root node to the selected node, which informs node selection in the future.

	Finally, once the predefined computational budget is reached, the root's child with the highest averaged reward will be chosen to indicate the corresponding question $\left(a_i ? a_j\right)$ to be posed in the $t$-th round.
	To sum up, the questioning policy $\pi \left(Q^{(t-1)}\right)$ is a decision-time planning algorithm by focusing more on the current state $Q^{(t-1)}$.
	It spends the computational budget to successively direct Monte Carlo simulations toward more highly rewarding trajectories and incrementally grows a lookup table to store each node's reward.
	Thus, the questioning policy $\pi \left(Q^{(t-1)}\right)$ avoids the problem of globally approximating a reward function while retaining the benefit of using past experience to guide exploration.
	\begin{footnotesize}
		\begin{breakablealgorithm} \label{alg:1}
			\caption{QUESTIONING POLICY $\pi \left(Q^{(t-1)}\right)$}
			\begin{algorithmic}[1] 
				\Function {$\pi \left(Q^{(t-1)}\right)$}{}
				\State create root node $node_0$ with state $Q^{(t-1)}$
				\State $N(node_0) \leftarrow 0$
				\State $V(node_0) \leftarrow 0$
				\While {within computational budget}
				\State $node \leftarrow$ SELECTION\&EXPANSION($node_0$)
				\State $\Delta \leftarrow$ SIMULATION($node$)
				\State BACKPROPAGATION($node, \Delta$)
				\EndWhile
				\State \Return question $\left(a_i ? a_j\right)$ corresponding to node $\text{BESTCHILD}(node_0, 0)$
				\EndFunction
				\Function {SELECTION\&EXPANSION}{$node$}
				\While {$node$ is non-terminal}
				\If {$node$ is not fully expanded}
				\State choose candidate question $\left(a_i ? a_j\right)$ that can be posed
				\State add question $\left(a_i ? a_j\right)$ as new child $node^\prime$ to $node$
				\State $N(node^\prime) \leftarrow 0$
				\State $V(node^\prime) \leftarrow 0$
				\State \Return $node^\prime$
				\Else
				\State $node \leftarrow \text{BESTCHILD}(node, C_p)$
				\EndIf
				\EndWhile
				\State \Return $node$
				\EndFunction
				\Function {BESTCHILD}{$node$, $c$}
				\State \Return $\argmax_{node^\prime\in \text{children of }node} \frac{V(node^\prime)}{N(node^\prime)}+2c\sqrt{\frac{2\ln N(node_0)}{N(node^\prime)}}$
				\EndFunction
				\Function {SIMULATION}{$node$}
				\State Randomly choose candidate questions to form a sequence of $T$ questions
				\State Predict participant's preferences for unanswered questions
				\State Calculate reward as variance reduction scale after posing the sequence of $T$ questions
				\State \Return reward for $node$
				\EndFunction
				\Function {BACKPROPAGATION}{$node, \Delta$}
				\While {$node$ is not null}
				\State $N(node) \leftarrow N(node)+1$
				\State $V(node) \leftarrow V(node)+\Delta$
				\State $node \leftarrow \text{parent of }node$
				\EndWhile
				\EndFunction
			\end{algorithmic}
		\end{breakablealgorithm}
	\end{footnotesize}

	\section{Approximating the Value Function Model $U(\cdot)$ Using Piecewise-Linear Functions} \label{app-D}
	In this section, we delve into the details of approximating the value function model $U(\cdot)$ with piecewise-linear marginal value functions. 
    Let $X_j = [\alpha_j, \beta_j]$ denote the performance scale on criterion $g_j$, where $\alpha_j$ and $\beta_j$ represent the worst and best performances of all the alternatives in $A$ on criterion $g_j$, respectively. 
    To build a piecewise-linear marginal value function $u_j(\cdot)$, the interval $X_j = [\alpha_j, \beta_j]$ is divided into $\gamma_j$ equal-length sub-intervals, represented as $[x_j^0, x_j^1], [x_j^1, x_j^2], \ldots, [x_j^{\gamma_j-1}, x_j^{\gamma_j}]$, where $x_j^{k}=\alpha_j+\frac{k}{\gamma_j}\left(\beta_j-\alpha_j\right)$, $k =
    0,1,\ldots,\gamma_j$.
    Then, the marginal value of alternative $a_i$ on criterion $g_j$ can be approximated through linear interpolation:
    
        \begin{equation*}
    \begin{aligned}
    u_{j}\left(g_j\left(a_i\right)\right)= & u_{j}\left(x_j^{k_j}\right)+\frac{g_j\left(a_i\right)-x_j^{k_j}}{x_j^{k_j+1}-x_j^{k_j}}\left(u_{j}\left(x_j^{k_j+1}\right)-u_{j}\left(x_j^{k_j}\right)\right), \\
    & \text { for } g_j\left(a_i\right) \in\left[x_j^{k_j}, x_j^{k_j+1}\right].
    \end{aligned}
    \end{equation*}
    
    According to the above formula, once the marginal values at knots (i.e., $u_{j}\left(x_j^0\right)$, $u_{j}\left(x_j^1\right)$, $\ldots$, $u_{j}\left(x_j^{\gamma_j}\right)$) are determined, the marginal value function $u_{j}(\cdot)$ can be fully specified. 
    
    Let $\Delta u_{j}^k=u_{j}\left(x_j^k\right)-u_{j}\left(x_j^{k-1}\right)$ denote the incremental difference between marginal values, $k = 1,\ldots,\gamma_j$. 
    Then, the marginal value $u_{j}(g_j(a))$ can be reformulated as:
    \begin{equation*}
    \begin{aligned}
    u_{j}\left(g_j\left(a_i\right)\right)= & \sum_{k=1}^{k_j} \Delta u_{j}^k+\frac{g_j\left(a_i\right)-x_j^{k_j}}{x_j^{k_j+1}-x_j^{k_j}} \Delta u_{j}^{k_j+1}, \\
    & \text { for } g_j\left(a_i\right) \in\left[x_j^{k_j}, x_j^{k_j+1}\right] .
    \end{aligned}
    \end{equation*}
    To facilitate the following parameter inference, we can represent the marginal value $u_{j}(g_j(a))$ by the inner product of two vectors $\Delta \bm{u}_{j} = \left(\Delta u_{j}^1, \ldots, \Delta u_{j}^{\gamma_j}\right)^\mathrm{T}$ and $\bm{V}_j(a) = \left(v_j^1\left(a_i\right), \ldots, v_j^{\gamma_j}\left(a_i\right)\right)^\mathrm{T}$ as $  u_{j}\left(g_j\left(a_i\right)\right)= \Delta \bm{u}_{j}^\mathrm{T} \bm{V}_j\left(a_i\right)$, where $v_j^k\left(a_i\right)$ for each $k =
    1, . . ., \gamma_j$ is defined as
    \begin{equation*}
    v_j^k\left(a_i\right)=\left\{\begin{array}{cl}
    1, & \text { if } g_j\left(a_i\right)>x_j^k, \\
    \frac{g_j\left(a_i\right)-x_j^{k-1}}{x_j^k-x_j^{k-1}}, & \text { if } x_j^{k-1} \leqslant g_j\left(a_i\right) \leqslant x_j^k, \\
    0, & \text { otherwise. }
    \end{array}\right.
    \end{equation*}
    Furthermore, let $\bm{u}=\left(\Delta \bm{u}_{1}^\mathrm{T}, \ldots, \Delta \bm{u}_{m}^\mathrm{T}\right)^\mathrm{T} \in \mathbb{R}^\gamma$ and $ \bm{V}\left(a_i\right)=$ $\left(\bm{V}_1\left(a_i\right)^\mathrm{T}, \ldots, \bm{V}_m\left(a_i\right)^\mathrm{T}\right)^\mathrm{T} \in \mathbb{R}^\gamma$, such that $\gamma = \sum\limits_{j=1}^m \gamma_j$ is the dimensionality. 
    Then, the comprehensive value of alternative $a_i$ can be expressed as
    \begin{equation*}
    U\left(a_i\right)=\bm{u}^\mathrm{T} \mathbf{V}\left(a_i\right) .
    \end{equation*}
    Note that $\bm{u}$ is a parameter vector to signify the intrinsic character of the preference model $U(\cdot)$, and $\bm{V}\left(a_i\right)$ is a characteristic vector that is fully determined by the performances of alternative $a_i$ on multiple criteria, particularly irrelevant to the DM's preferences. 
    In other words, the DM's preference model $U(\cdot)$ is parameterized by the preference vector $\bm{u}$, and the following parameter inference procedure only needs to infer the vector $\bm{u}$.
    Moreover, to ensure the monotonicity of the marginal value functions $u_{j}(\cdot)$ and normalize the comprehensive
    value $U(\cdot)$ within the interval [0,1], it's essential to account for the following linear constraints:
    \begin{equation*}
    \left\{\begin{array}{l}
    \bm{u} \geqslant \bm{0}, \\
    \bm{1} \cdot \bm{u}= 1,
    \end{array}\right.
    \end{equation*}
    where $\bm{1}$ and $\bm{0}$ are the vectors with all entries being equal to 1 and 0, respectively.
    
    The advantages of using piecewise-linear functions for approximating the DM's true preference model consist of the following aspects. First, given a sufficient number of sub-intervals, a piecewise-linear function can approximate any form of a non-linear function~(\cite{liu2021data}). 
    This property enhances the flexibility and the expressiveness of such a preference model in representing the DM's sophisticated preferences~(\cite{kadzinski2017expressiveness}).
    Moreover, the shape of a piecewise-linear marginal value function (e.g., linear, concave, or convex) reflects the DM's risk attitude (e.g., neutral, risk-averse, or risk-seeking)~(\cite{fishburn1979two,ghaderi2021incorporating}). 
    Also, using such functions is computationally tractable in a probabilistic inference procedure. 
    Note that the number of sub-intervals on each criterion $\gamma_j$ are hyperparameters and can be specified using cross-validation.

	\section{Experimental Design for Preference Inference Experiments}\label{app-E}

	In this section, we describe the detailed experimental design for preference inference experiments. 
	We aim to assess the impact of both using the reparameterization trick that should enable more efficient optimization using gradient-based methods and avoiding arbitrary assumptions concerning the distribution in the space of preference models. Two variants of the proposed variational Bayesian inference approach are implemented: one with RT and the other without. We also compare these variants with the Stochastic Ordinal Regression (SOR) method (\cite{kadzinski2013robust}), which infers a preference model by exploring the polyhedron defined by linear constraints derived from preference information. SOR assumes a uniform distribution over the space of compatible preference models and uses Monte Carlo simulation to compute the proportion of model instances confirming one alternative's preference over another.

	The comparison between the proposed variational Bayesian approach and SOR is based on the \textit{Average Support of the inferred Pairwise Outranking Indices (ASP)}. ASP is defined as:

	\begin{small} \begin{equation} ASP = \frac{2}{n(n - 1)}\sum_{i = 1}^n\sum_{\substack{j = 1 \ j \neq i}}^n POI(a_i, a_j) \cdot \mathbb{I}(U^{true}(a_i) \geqslant U^{true}(a_j)), \end{equation} \end{small} \noindent where $POI(a_i, a_j)$ is the share of sampled value functions where $a_i$ is not worse than $a_j$, and $\mathbb{I}(U^{true}(a_i) \geqslant U^{true}(a_j))$ equals 1 if $U^{true}(a_i) \geqslant U^{true}(a_j)$, and 0 otherwise. ASP measures the agreement between the true pairwise comparisons and those inferred by the method. The variational Bayesian approach and SOR sample from the posterior distribution and the polyhedron defined by linear constraints, respectively. ASP ranges from 0 to 1, with higher values indicating better performance.

	The real-world datasets used in this study are drawn from prior MCDA research in fields such as education, medicine, industry, and marketing, with sizes ranging from 14 to 118 alternatives and 3 to 8 criteria. A summary of these datasets is provided in Table \ref{tab:datastes}. The experimental analysis is based on random data generated through Monte Carlo simulations, considering three factors: the shape of the marginal value function ($F_1$), the number of pairwise comparisons provided by the DM ($F_2$), and the proportion of biased preference information ($F_3$). These factors are closely linked to the inference of the preference model. The levels of these factors in the simulation are detailed in Table~\ref{tab:factors}.

	\begin{table} 
		\caption{The characteristics of eight real-world datasets.} 
		\centering\footnotesize
		\label{tab:datastes}
		\begin{tabularx}{15.4cm}{llcc | llcc} 
			\hline 
			No. & Dataset & Alt. No. & Crit. No. & No. & Dataset & Alt. No. & Crit. No.\\ 
			\hline 
			1 & Thierrys'choice (TC) & 14 & 5 & 5 & Research units (RU) & 93 & 4 \\
			2 & Industry competitiveness (IC) & 31 & 6 & 6 & EIU best cities (EBC) & 70 & 7\\
			3 & Student selection (SS) & 76 & 6 & 7 & potential index (MPI) & 86 & 8\\
			4 & Couple's embryos (CE) & 51 & 7 & 8 & Guardian UK universities (GUU) & 118 & 8 \\
			\hline
		\end{tabularx} 
	\end{table}
	\begin{table} 
		\caption{Levels for the considered factors in the simulation.} 
		\centering\footnotesize
		\label{tab:factors}
		\begin{tabularx}{14cm}{ll} 
			\hline 
			Factors & Levels\\ 
			\hline 
			$F_1$: the shape of marginal value function & linear, concave, convex, mixture \\ 
			$F_2$: the number of pairwise comparisons & 20, 40, 60, 80 \\ 
			$F_3$: the proportion of biased preference information & 0, 10\%, 20\%, 30\%\\
			\hline 
		\end{tabularx} 
	\end{table} 

	To obtain insights based on the three factors delineated in Table \ref{tab:factors}, we assume the true preference models of 20 DMs to generate the corresponding preference information for each combination of the factors presented in Table \ref{tab:factors} (64 combinations). We adopt the procedure in \cite{ghaderi2021incorporating} to construct a value function $U^{true}(a) = \sum_{j=1}^m u_j^{true}(g_j(a))$ serving as the DM's true preference model. Initiating this process requires extracting a criteria weight vector, $\bm{w} = (w_1, \ldots, w_m)^T$, from a Dirichlet distribution characterized by the parameter vector whose all entries are equal to one. This step guarantees that each criterion weight $w_j$ is positive and their total sum equals one. Subsequently, the simulation of each criterion's marginal value function $u_j^{true}(\cdot)$, distinguished by its shape (F1), adheres to the following paradigm:
	\begin{equation}
		u_j^{true}(x)=
		\begin{cases} 
		  w_j x, & \text{for a linear} \ u_j^{true}(\cdot), \\
		  w_j \frac{1-\exp(-c_jx)}{1-\exp(-c_j)}, & \text{for a concave or convex} \ u_j^{true}(\cdot). 
	   \end{cases}
	\end{equation}
	
	The parameter $c_j$, nonzero and randomly sampled from the interval $[-10, 10]$ under a uniform distribution, governs the curvature of $u_j^{true}(\cdot)$. A positive (negative) $c_j$ indicates a concave (convex) form for $u_j^{true}(\cdot)$, reflecting the DM's risk attitude concerning criterion $g_j$, in accordance with the insights provided by  \cite{ghaderi2021incorporating}. Regarding factor $F_1$, we contemplate four potential shapes: linear, concave, convex, and a hybrid, which we denominate as a mixture. The mixture involves a random selection among the other three forms for constructing the marginal value functions.
	
	Then, the assumed true preference model of the DM, $U^{true}(\cdot)$, is employed to generate the preference information. Specifically, we utilize the true value function $U^{true}(a)$ to calculate a true comprehensive value for each alternative $a \in A$. Subsequently, we select a predefined number of pairwise comparisons ($F_2$) in accordance with the true preference relations among alternatives to form the preference information. In the experiment, a certain proportion ($F_3$) of biased preference information is also taken into consideration. To generate such data, we sort all potential pairs of alternatives based on the difference between their true comprehensive values as indicated by $U^{true}(\cdot)$. We then select a certain proportion (0\%, 10\%, 20\%, 30\%) of pairs with the smallest value differences and invert their preference relations. This setting is reasonable, as DMs in real situations are more likely to provide inverse pairwise comparisons for pairs of alternatives with negligible value differences. Bias introduced by inverting the true preference relations which may result in no value function being fully compatible with all comparisons. Consequently, SOR becomes inapplicable in its standard form. To address this, we incorporate the inconsistency resolution method from \cite{mousseau2003resolving}, which identifies and removes constraints likely causing inconsistencies. In contrast, our Bayesian approach, which assigns probabilities to all possible preference models, inherently accommodates such inconsistencies without modification.

	In the proposed variational Bayesian inference approach with/without RT and SOR, two key parameters must be set: (1) the number of sub-intervals $\gamma_j$ for each criterion $g_j$ $(j = 1, \dots, m)$, and (2) the number of samples $W$ for approximating the posterior distribution in the variational Bayesian approach. For simplicity, we assume all criteria have the same number of sub-intervals. We determine the optimal $\gamma_j$ using the Deviance Information Criterion (DIC) (\cite{gelman2014understanding}), with values {1, 2, 3, 4, 5}. DIC balances model performance and complexity, which is commonly used in Bayesian model selection. We set $W = 10,000$, as this is sufficient for posterior approximation (\cite{tervonen2007implementing}). In the variational Bayesian method, additional parameters are also necessary to be determined: the maximum number of iterations $T$, the number of samples $W_g$ for calculating the stochastic gradient, and the hyperparameters for the Adam optimizer (learning rate $\eta$, exponential decay rates for moment estimates $\beta_1$ and $\beta_2$, and a small constant $\epsilon$ to avoid division by zero). Based on experimental testing, we set the following values to balance inference accuracy with computational cost: $T = 500$, $W_g = 10,000$, $\eta = 0.01$, $\beta_1 = 0.9$, $\beta_2 = 0.999$, and $\epsilon = 10^{-8}$. We implement SOR using R's open-source package \emph{hit-and-run} (\cite{tervonen2007implementing}). 

	\section{Wilcoxon Test for Preference Inference Experiments} \label{app-F}
	The results in Section 4 show that the RT-enabled variant consistently achieves higher ASP values than both its non-RT counterpart and the SOR method. To assess the statistical significance of these findings, we conducted a Wilcoxon test comparing the ASP values of the proposed variational Bayesian inference approach with RT against the other two methods. As presented in Table \ref{tab:Wilcoxontest}, the $p$-values for nearly all datasets are below $0.001$, confirming that our RT-based approach significantly outperforms the alternatives.
	\begin{table}[ht]
		\centering\footnotesize
		\caption{Wilcoxon Test on ASP for the Proposed Variational Bayesian Inference Approach with RT vs. Other Two Methods}
		\label{tab:Wilcoxontest}
		\begin{tabular}{llcccccccc}
			\toprule
			\multicolumn{2}{c}{\multirow{2}{*}{\makecell[c]{\textbf{Method}                                                                                    \\ \textbf{Comparison}}}} & \multicolumn{8}{c}{\textbf{Dataset}} \\
			\cmidrule(lr){3-10}
												   &           & TC        & IC        & SS        & CE        & RU        & EBC       & MPI       & GUU       \\
			\midrule
			\multirow{2}{*}{\textbf{RT vs. No RT}} & statistic & 895,340   & 848,923   & 915,079   & 967,135   & 956,578   & 926,460   & 910,440   & 921,789   \\
												   & $p$-value & 4.665e-05 & 1.119e-01 & 2.938e-07 & 2.548e-15 & 2.031e-13 & 9.690e-09 & 1.065e-06 & 4.105e-08 \\
			\midrule
			\multirow{2}{*}{\textbf{RT vs. SOR}}   & statistic & 947,991   & 1,006,954 & 976,260   & 925,663   & 956,478   & 886,898   & 903,802   & 911,814   \\
												   & $p$-value & 5.679e-12 & 1.010e-23 & 4.494e-17 & 1.245e-08 & 2.115e-13 & 2.942e-04 & 6.059e-06 & 7.315e-07 \\
			\bottomrule
		\end{tabular}
	\end{table}

	\section{Comprehensive Computational Results Across Different Datasets for Preference Inference Experiments} \label{app-G}
	In this section, we provide a detailed presentation of the computational results across different datasets and experimental settings for the preference inference experiments.

	\subsection{Performance analysis for different shapes of marginal value functions} 

	In this section, we provide a detailed analysis of the experiments considering different shapes of marginal value functions across various datasets. Figure \ref{fig:ASPresults_F1} presents boxplots for the average ASP values, characterizing the ability to infer DMs' preferences using the variants of the proposed variational Bayesian inference approach with/without RT and the SOR method. Note that the solid line represents the median of the dataset, while the dashed line indicates the mean.
	\begin{figure}[!htbp]
		\centering 
		\includegraphics[scale=0.4]{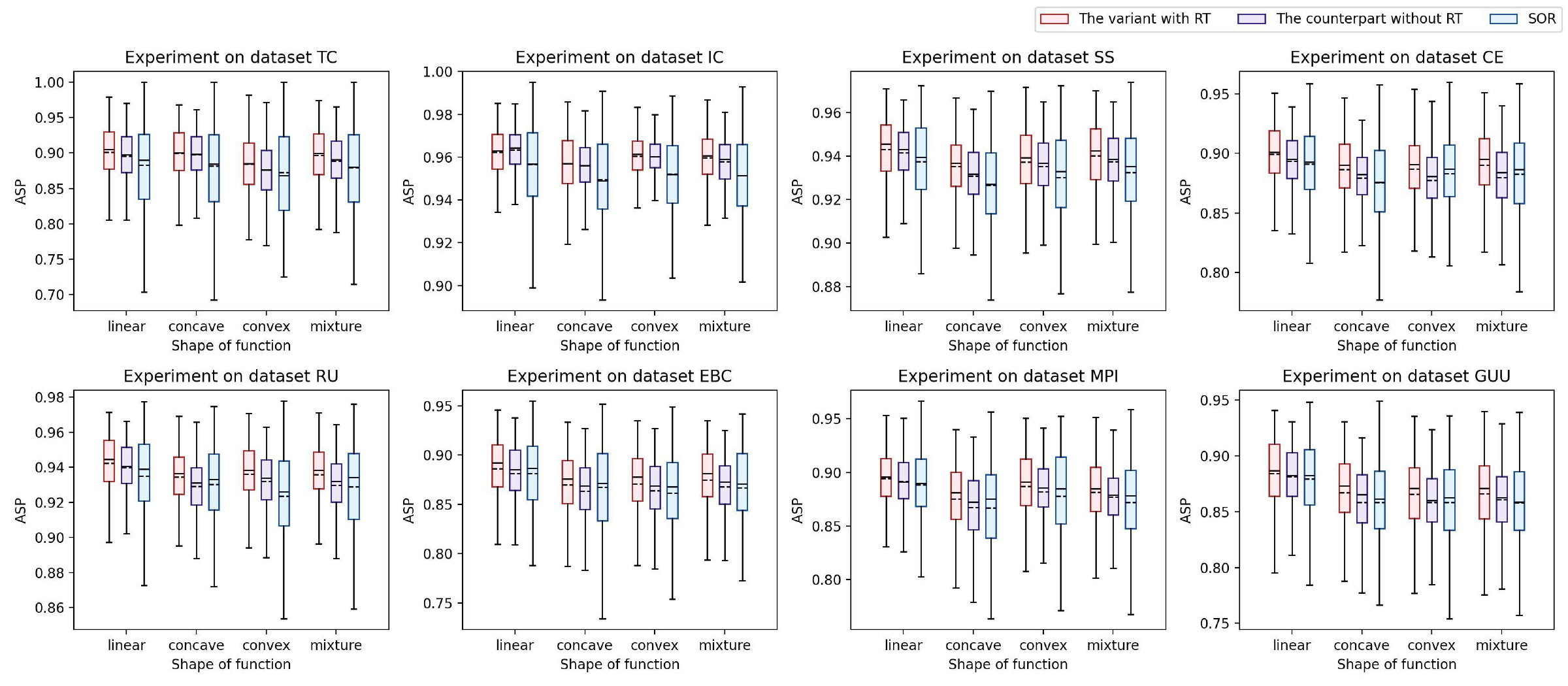}
		\caption{Average ASP for the variants of the proposed variational Bayesian inference approach with/without RT and SOR, considering different shapes of marginal value functions across various datasets.} \label{fig:ASPresults_F1}
	\end{figure}

	By examining the mean and median values, it is evident that the proposed method consistently outperforms the other two methods regardless of the form of the marginal value function. This advantage remains across all four types of marginal value functions. Additionally, the figures demonstrate that the results obtained by the two variants of the proposed variational Bayesian inference approach are more concentrated, indicating a more stable performance in inferring DMs' preferences. Finally, the ASP values for all three preference inference methods are higher when the marginal value function is linear compared to the other three non-linear forms. This trend is reasonable because we approximate the true marginal value function using the piecewise linear form. Simpler functions require fewer sub-intervals and parameters, making them easier to estimate. Conversely, more complex functions necessitate more sub-intervals and parameters for accurate approximation, making the estimation process more challenging.

	\subsection{Performance analysis for different numbers of pairwise comparisons}

	This section discusses the results of experiments involving varying amounts of preference information. We compare the variants of the proposed variational Bayesian inference approach with/without RT and SOR, given $\{20, 40, 60, 80\}$ pairwise comparisons. As shown in Figure \ref{fig:ASPresults_F2}, with the increase in the number of pairwise comparisons, there is a gradual improvement in the performance of all three methods. Additional preference information reduces the uncertainty in inferring the DM's preference model. The performance of SOR relies more heavily on the amount of preference information compared to our variational Bayesian methods. This disparity stems from the intrinsic characteristics of the two approaches. SOR translates the given preference information into linear constraints and explores a set of compatible value functions within the feasible region. Consequently, additional preference information reduces the volume of the feasible region, thereby decreasing the uncertainty in inferring the DM's preference model. In contrast, our approach models the given preference information from a probabilistic perspective and accounts for the differences between instances of value functions. As a result, even with a small amount of preference information, our method infers a concentrated posterior distribution of the value function, reducing the uncertainty in exploring the DM's preferences.

	However, as shown in Figure \ref{fig:ASPresults_F2}, the ASP values derived by the SOR method actually decrease when the amount of preference information increases from 60 to 80 for the first two datasets (TC and IC). This abnormal result arises because, as the number of preference information increases, the amount of biased preference inputs also proportionally increases. The SOR method, which infers the preference model based on feasible regions defined by linear constraints constructed from preference information, is significantly affected by biased preference information as it can exclude the true preference model from the feasible region. In contrast, our method models preferences from a probabilistic perspective, assigning probabilities to all possible preference models and thus preventing the direct exclusion of the true preference model due to biased information. Appendix \ref{app-G3} provides more direct evidence for this point. Moreover, as the amount of accurate preference information increases, it helps to correct the impact of biased data, which is consistent with the conclusions drawn from Figure \ref{fig:ASPresults_F2}.
	\begin{figure}[!htbp]
		\centering
		\includegraphics[scale=0.4]{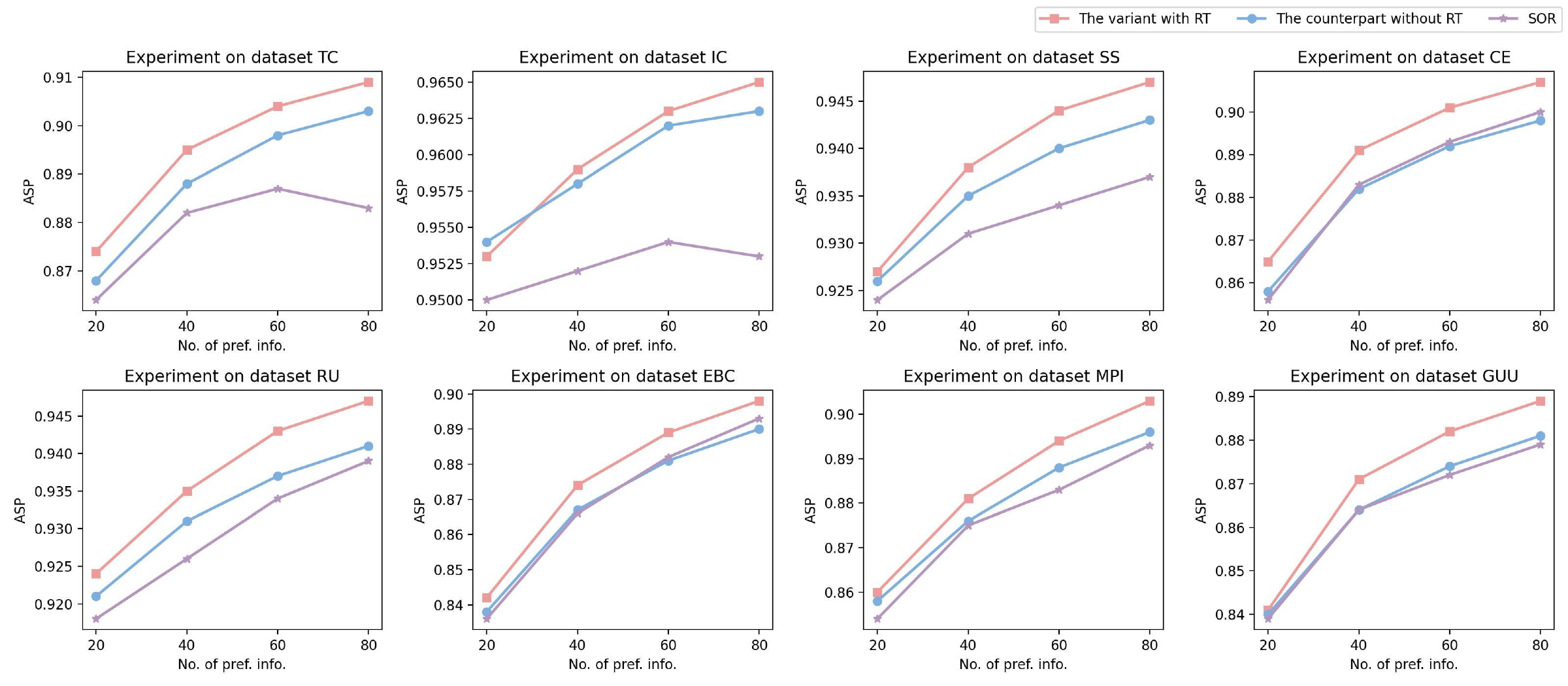}
		\caption{Average ASP for the variants of the proposed variational Bayesian inference approach with/without RT and SOR, considering different numbers of pairwise comparisons across various datasets.} \label{fig:ASPresults_F2}
	\end{figure}

	\subsection{Performance analysis for different proportions of biased preference information} \label{app-G3}

	In this section, we examine the performance of the variants of the proposed variational Bayesian inference approach with/without RT and SOR with biased preference information. 
	Figure \ref{fig:ASPresults_F3} summarizes the average ASP for the three methods considering different proportions of biased preference information across various datasets. The observations can be summarized from the following two aspects. On the one hand, as the proportion of biased preference information increases, the performance of all three methods declines, which is in line with our expectations. This is because an increase in biased preference information causes the posterior distribution of the value function to diverge from the true preference model $U^{true}(a)$ for the proposed Bayesian methods. As for SOR, it causes the feasible region of the preference model to be wrongly restricted. On the other hand, the magnitude of the decline in ASP with increasing biased preference information is less significant with our approach compared to SOR, indicating that the proposed variational Bayesian approach exhibits greater robustness to biased preference information. The reason is that our method assigns probabilities to all potential value function models, ensuring that the actual DM's preference model remains within the consideration set despite the presence of biased information. Moreover, by modeling the provided preferences from a probabilistic perspective, our approach ensures that even if a small proportion of comparisons is biased, the majority of consistent preference information will guide the posterior distribution of the value function closer to the assumed true preference model (\cite{ru2022bayesian}).

	\begin{figure}[!htbp]
		\centering
		\includegraphics[scale=0.4]{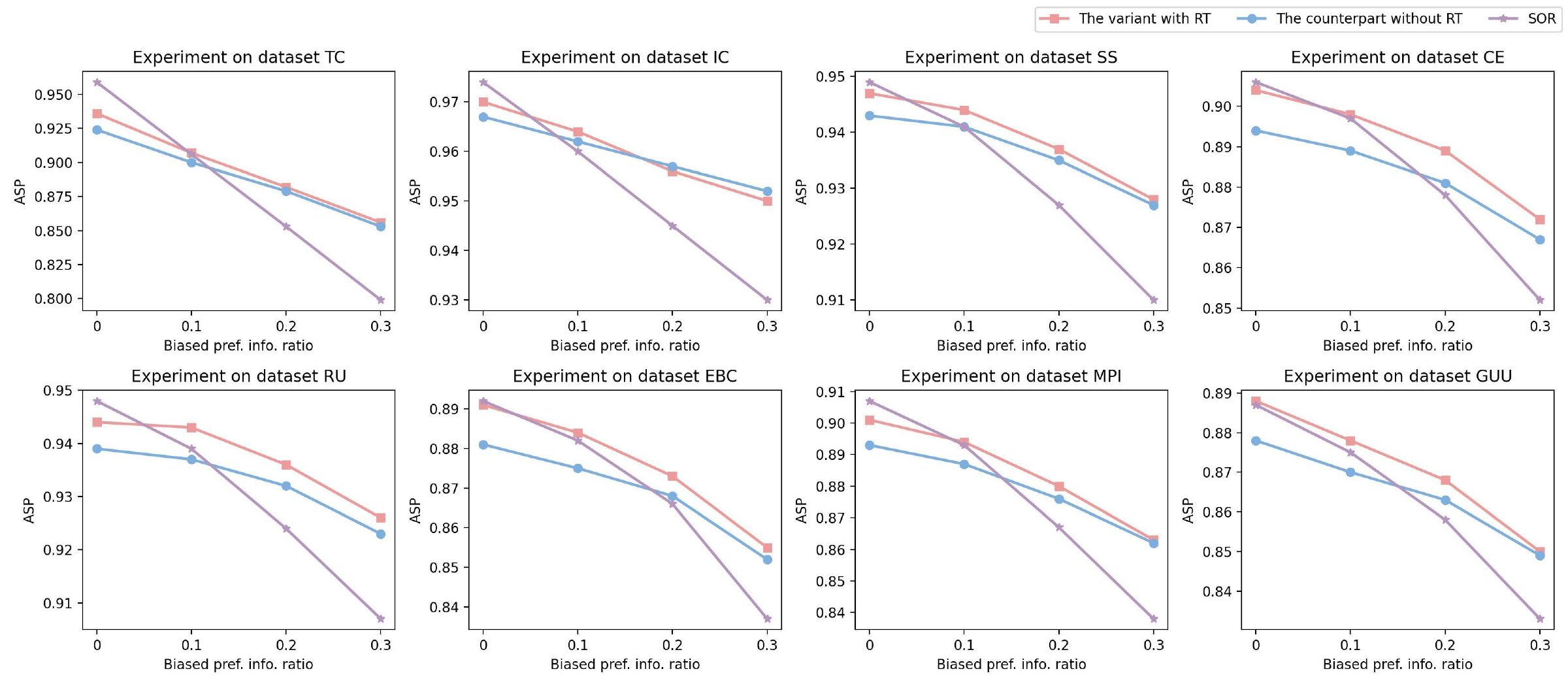}
		\caption{Average ASP for the variants of the proposed variational Bayesian inference approach with/without RT and SOR, considering different proportions of biased preference information across various datasets.} \label{fig:ASPresults_F3}
	\end{figure}

	\subsection{Comparison of the variants of the proposed variational Bayesian inference approach with/without RT in high variance problem} \label{sec:Performance_RT}

	In this section, we compare the performance of the variants of the proposed variational Bayesian inference approach with/without RT and analyze the underlying reasons. Figure \ref{fig:ASPresults_F1}, \ref{fig:ASPresults_F2}, and \ref{fig:ASPresults_F3} illustrate that the ASP values obtained by the two variants exhibit similar trends across different problem settings ($F_1$, $F_2$, $F_3$). However, the variant with RT consistently outperforms the counterpart without RT. This aligns with our expectation, as the counterpart without RT is susceptible to high variance issues, leading to significant instability
	in gradient computation. To further investigate the effectiveness of RT in mitigating high variance, we conducted an additional experiment. In this experiment, we performed gradient computations 100 times for each iteration of gradient ascent for both variants. We then calculated the variance of these 100 gradients for each dimension. To present the results more intuitively, we averaged the variances across different numbers of iterations, problem settings, and dimensions. 

	The final averaged results for various datasets are shown in Figure \ref{fig:RT}. In the box plots, the dashed lines represent the mean values, while the solid lines indicate the medians. Note that since the results obtained from the variant with RT are substantially smaller than those from the counterpart without RT (for example, in the first dataset, the median for the results of the variant with RT is 0.02, compared to 7.35 for the counterpart without RT), we have used separate y-axes on the left and right sides of the plot to display both sets of results in one figure. From this outcome, it can be discerned that the gradient variance computed by the variant with RT is considerably lower than the counterpart without RT, thereby explaining the effect of the RT in enhancing the inference performance.
	\begin{figure}[!htbp]
		\centering
		\includegraphics[scale=0.5]{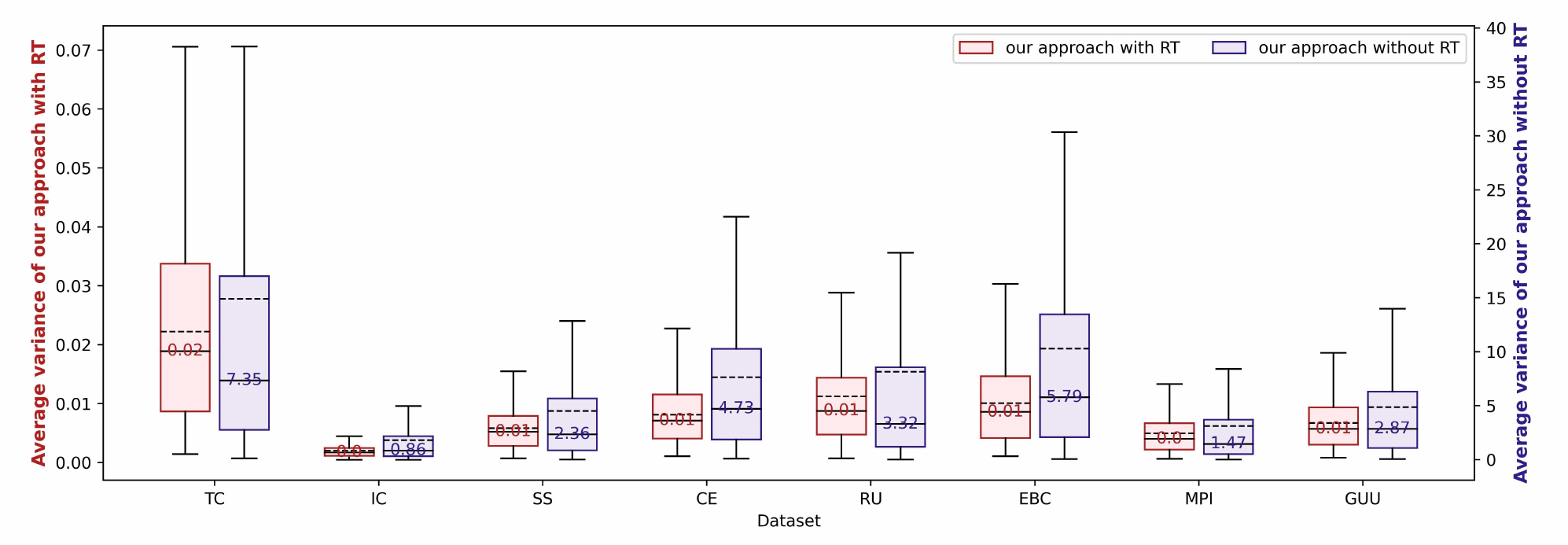}
		\caption{The gradient variance of the variant with RT and the counterpart without RT on each dataset.} \label{fig:RT}
	\end{figure}
	
	\section{Experimental Design for Questioning Policy Experiments} \label{app-H}

	In this section, we describe the detailed experimental design for questioning policy experiments. 
	
	We consider an iterative question-answer process for preference elicitation in MCDA. In this process, the proposed questioning policy based on MCTS is used to select the optimal question during each round of interaction, leveraging the preference information gathered from previous interactions. The objective of the questioning policy is to collect the most informative preference data from the DM within a constrained number of interaction rounds.

	We assess the sufficiency of the elicited preference information by measuring the uncertainty regarding the DM's preferences at the end of the questioning process. Lower uncertainty indicates more comprehensive preference information, while higher residual uncertainty suggests fewer insights. To quantify uncertainty reduction after multiple interactions, we use the following typical metrics based on the variational Bayesian approach introduced in Section 2.2:
	\begin{itemize}
		\item $f_\textit{VAR}$, which is the variance of the posterior distribution derived from the variant of the proposed variational approach with RT given the preference information obtained after a limited number of interactions.
		\item $f_\textit{PWI}$, which quantifies the uncertainty for all pairs of alternatives based on the given preference information obtained through the interactions and is built on the pairwise winning index (\textit{PWI}) (see \cite{ciomek2017heuristics}). The pairwise winning index $\textit{PWI}_Q(a_i, a_j)$ represents the share of value functions sampled from the posterior distribution derived through the variant of the proposed variational approach with RT, given the preference information $Q$, in which $a_i$ is preferred to $a_j$. As such, $\textit{PWI}_Q(a_i, a_j)$ reflects the probability of $a_i \succ a_j$ given preference information $Q$. Consequently, for any $(a_i, a_j)\in A \times A$: $\textit{PWI}_Q(a_i, a_j)\in [0, 1]$ and $\textit{PWI}_Q(a_i, a_j) + \textit{PWI}_Q(a_j, a_i) \le 1$. The index $\textit{PWI}_Q(a_i, a_j)$ is usually estimated using Monte Carlo simulation (\cite{van2014notes}). Then, the metric $f_\textit{PWI}$ can be computed through the concept of Shannon entropy:
		\begin{small}
		\begin{equation}
			f_\textit{PWI}(Q)=\sum_{(a_i, a_j)\in A\times A, i\neq j} \frac{-\textit{PWI}_Q (a_i, a_j)\log_2 \textit{PWI}_Q (a_i, a_j)}{n\cdot (n-1)}.
		\end{equation}	
		\end{small}

		\item $f_\textit{RAI}$, which quantifies the uncertainty in the ranks attained by all alternatives based on the preference information obtained through the interactions and is built on the rank acceptability index ($\textit{RAI}$) (see \cite{ciomek2017heuristics}). The rank acceptability index $\textit{RAI}_Q(a_i, k) \in [0, 1]$ represents the proportion of value functions sampled from the posterior distribution derived through the variant of the proposed variational approach with RT given the preference information $Q$, that assign alternative $a$ the rank $k$. Similar to the pairwise winning index, $\textit{RAI}_Q(a_i, k)$ is usually estimated using Monte Carlo simulation, reflecting the probability that $a_i$ attains the $k$-th rank. For each $a_i \in A$, $\sum_{k=1}^n \textit{RAI}_Q(a_i, k) = 1$ holds. Then, the measure $f_\textit{RAI}$ can be computed through the concept of entropy:
		\begin{equation}
			f_{\textit{RAI}}(Q)=\sum_{i=1}^n \sum_{k=1}^n \frac{-\textit{RAI}_Q(a_i, k)\log_2 \textit{RAI}_Q(a_i, k)}{n}.
		\end{equation}

		\end{itemize}

	We compared the MCTS-based questioning policy in our approach with those from existing studies (\cite{ciomek2017heuristics}): two heuristics for selecting the myopically optimal pairwise question ($H_\textit{PWI}$ and $H_\textit{RAI}$), their extensions that utilize greater search depths ($H_{\textit{PWI}-2}$ and $H_{\textit{RAI}-2}$), the benchmark method $H_\textit{DVF}$, and a random selection policy, denoted as $H_\textit{RAND}$. Note that for all questioning policies, the next question must be selected from those that have not been previously asked, which we refer to as candidate questions. The details of these comparative questioning policies are described as follows:
	\begin{itemize}
		\item $H_\textit{PWI}$ and $H_\textit{RAI}$ choose the question that maximizes the estimated reduction in pairwise preference entropy and rank-related entropy, respectively, given preference information $Q$ as the next best question, which can be formulated as follows:
	\begin{equation}
		\begin{aligned}
			H_{\textit{X}}(Q) = 
			& \argmax_{(a_i ? a_j)\in \text{Candidate Questions}} (\frac{\textit{PWI}_Q(a_i, a_j)(f_\textit{X}(Q^{(t-1)})-f_\textit{X}(Q \cup \{a_i, a_j\}))}{\textit{PWI}_{Q}(a_i, a_j)+\textit{PWI}_Q(a_j, a_i)}+\\
			& \frac{\textit{PWI}_Q (a_j, a_i)(f_\textit{X}(Q)-f_\textit{X}(Q \cup \{a_i, a_j\}))}{\textit{PWI}_Q(a_i, a_j)+\textit{PWI}_Q(a_j, a_i)}),
		\end{aligned}
	\end{equation}
		where $X \in \{\textit{PWI}, \textit{RAI}\}$.
		\item $H_{\textit{PWI}-d}$ and $H_{\textit{RAI}-d}$ extent $H_\textit{PWI}$ and $H_\textit{RAI}$ to the greater search depths, which alleviates the problem of shortsightedness to some extent. The generalization of a heuristic $H_{X-d}$ to use search depth $d$ can be formulated as follows:
		\begin{equation}
			\begin{aligned}
				H_{\textit{X}-d}(Q) = 
				& \argmax_{(a_i ? a_j)\in \text{Candidate Questions}} (\frac{\textit{PWI}_Q(a_i, a_j) v(Q \cup \{a_i, a_j\}, \tilde{f}_\textit{X}, d-1)}{\textit{PWI}_{Q}(a_i, a_j)+\textit{PWI}_Q(a_j, a_i)}+\\
				& \frac{\textit{PWI}_Q(a_j, a_i) v(Q \cup \{a_j, a_i\}, \tilde{f}_\textit{X}, d-1)}{\textit{PWI}_Q(a_i, a_j)+\textit{PWI}_Q(a_j, a_i)}),
			\end{aligned}
		\end{equation}

		where  
		\begin{equation} v(Q, \tilde{f}_X, t)=\left\{
		\begin{aligned}
		&-\tilde{f}_X(Q),\quad \text{if} \ t=0, \\
		&\max_{(a_i ? a_j)\in \text{Candidate Questions}}
		(\frac{\textit{PWI}_Q(a_i, a_j) v(Q \cup \{a_i, a_j\}, \tilde{f}_\textit{X}, t-1)}{\textit{PWI}_{Q}(a_i, a_j)+\textit{PWI}_Q(a_j, a_i)}+ \\
		&\frac{\textit{PWI}_Q(a_j, a_i) v(Q \cup \{a_j, a_i\}, \tilde{f}_\textit{X}, t-1)}{\textit{PWI}_Q(a_i, a_j)+\textit{PWI}_Q(a_j, a_i)}),\quad \text{otherwise},
		\end{aligned}
		\right.
	\end{equation}

		and $X \in \{\textit{PWI}, \textit{RAI}\}$. Considering the factor of computational efficiency, we only compared the case where $d=2$.
		
		\item $H_\textit{DVF}$ accounts for the distribution of compatible value functions, splitting their polyhedron as evenly as possible based on $\textit{PWIs}$. This corresponds to choosing the question $\{a_i ? a_j\}$ for which the minimal pairwise winning index out of $\textit{PWI}_Q(a_i, a_j)$ and $\textit{PWI}_Q(a_j, a_i)$ is maximal:
		\begin{equation}
			H_\textit{DVF}(Q) = \argmax_{(a_i ? a_j)\in \text{Candidate Questions}} \min\{\textit{PWI}_Q(a_i, a_j), \textit{PWI}_Q(a_j, a_i)\}.
		\end{equation}

		\item $H_\textit{RAND}$ randomly selects the next interactive question from the pool of candidate questions.

	\end{itemize}

	\noindent Then, we compare the MCTS-based questioning policy in our approach with $H_{X-d}$ for $X \in \{\textit{PWI}, \textit{RAI}\}$, $d \in \{1, 2\}$, $H_\textit{DVF}$ and $H_\textit{RAND}$ in terms of above three metrics for the following problem instances: 
	\begin{itemize}
	\item $6-10$ alternatives with 3 criteria;
	\item $2-5$ criteria with 8 alternatives.
	\end{itemize}
	For each problem setting, we construct 20 instances by drawing the performance of alternatives on each criterion from a uniform distribution in [0, 1]. A problem instance for each dataset is created using the procedure in Appendix E to randomly generate a value function model $U^{true}(a) = \sum_{j=1}^m u_j^{true}(g_j(a))$ as the DM's preference model. For simplicity, we assume all criteria have the same number of sub-intervals and use cross-validation to determine $\gamma$. In the MCTS-based questioning policy, we balance computational efficiency with adequate exploration of the question space. Experimental tests showed that a computation budget below 500 kept computation time reasonable, and increasing it beyond 300 did not significantly improve performance but added computational burden. Therefore, we set the budget at 300 for the experimental study. For each problem instance, we used these questioning policies to ask $s$ questions, $s \in {2, 4, 6, 8}$, resulting in performance analysis over 720 interactive processes.

	\section{Wilcoxon Test for Questioning Policy Experiments} \label{app-I}
	The results in Section 5 demonstrate that our MCTS-based questioning policy consistently achieves the lowest values across all three evaluation metrics ($f_\textit{VAR}$, $f_\textit{PWI}$, and $f_\textit{RAI}$) highlighting its effectiveness in reducing uncertainty in preference inference. Statistical validation through the one-sided Wilcoxon test (Table \ref{tab:one-sided Wilcoxon test}) further confirms the superiority of our approach. In contrast, the worst-performing method, $H_\textit{RAND}$, yields significantly higher metric values, emphasizing the critical role of active exploration strategies in efficient preference elicitation.

	\begin{table}[t]
	\centering \footnotesize
	\caption{Results ($p$-values) of the Wilcoxon signed-rank test for the significance of differences in the average values of $f_\textit{VAR}$, $f_\textit{PWI}$, and $f_\textit{RAI}$ between different pairs of questioning policies, based on the performance across $640$ instances (values denoted as $0.000^+$ indicate a small positive value less than $0.001$; the numbers in parentheses represent the number of instances where the questioning policy in the row yields a strictly lower value compared to the policy in the column).}
	\setlength{\abovecaptionskip}{0.05cm}
	\centering
	\label{tab:one-sided Wilcoxon test}
	\renewcommand\arraystretch{0.85}
	\resizebox{1\linewidth}{!}{
		\begin{tabular}{*{15}{c}}
			\toprule
			Metric                            & Questioning policy   & Our approach & $H_\textit{PWI}$ & $H_\textit{RAI}$ & $H_{\textit{PWI}-2}$ & $H_{\textit{RAI}-2}$ & $H_\textit{DVF}$ & $H_\textit{RAND}$ \\
			\midrule
			\multirow{7}{*}{$f_\textit{VAR}$}
			                                  & Our approach         & -            & $0.000^+$ (466)  & $0.000^+$ (477)  & $0.000^+$ (465)      & $0.000^+$ (471)      & $0.000^+$ (474)  & $0.000^+$ (588)   \\
			                                  & $H_\textit{PWI}$     & 1.000 (174)  & -                & $0.000^+$ (364)  & 0.089 (328)          & 0.001 (360)          & 0.001 (345)      & $0.000^+$ (544)   \\
			                                  & $H_\textit{RAI}$     & 1.000 (163)  & 1.000 (276)      & -                & 0.985 (290)          & 0.897 (295)          & 0.428 (309)      & $0.000^+$ (525)   \\
			                                  & $H_{\textit{PWI}-2}$ & 1.000 (175)  & 0.911 (312)      & 0.015 (350)      & -                    & 0.031 (341)          & 0.042 (336)      & $0.000^+$ (538)   \\
			                                  & $H_{\textit{RAI}-2}$ & 1.000 (169)  & 0.999 (280)      & 0.103 (345)      & 0.969 (299)          & -                    & 0.167 (332)      & $0.000^+$ (534)   \\
			                                  & $H_\textit{DVF}$     & 1.000 (163)  & 0.999 (295)      & 0.572 (331)      & 0.958 (304)          & 0.833 (308)          & -                & $0.000^+$ (521)   \\
			                                  & $H_\textit{RAND}$    & 1.000 (52)   & 1.000 (96)       & 1.000 (115)      & 1.000 (102)          & 1.000 (106)          & 1.000 (119)      & -                 \\
			\midrule
			\multirow{7}{*}{$f_\textit{PWI}$} & Our approach         & -            & $0.000^+$ (408)  & $0.000^+$ (430)  & $0.000^+$ (419)      & $0.000^+$ (429)      & $0.000^+$ (434)  & $0.000^+$ (524)   \\
			                                  & $H_\textit{PWI}$     & 1.000 (232)  & -                & $0.000^+$ (350)  & 0.009 (350)          & $0.000^+$ (359)      & $0.000^+$ (361)  & $0.000^+$ (485)   \\
			                                  & $H_\textit{RAI}$     & 1.000 (210)  & 1.000 (290)      & -                & 0.999 (290)          & 0.914 (295)          & 0.912 (298)      & $0.000^+$ (446)   \\
			                                  & $H_{\textit{PWI}-2}$ & 1.000 (221)  & 0.991 (290)      & 0.001 (350)      & -                    & 0.004 (349)          & 0.045 (339)      & $0.000^+$ (456)   \\
			                                  & $H_{\textit{RAI}-2}$ & 1.000 (211)  & 1.000 (281)      & 0.086 (345)      & 0.996 (291)          & -                    & 0.637 (313)      & $0.000^+$ (444)   \\
			                                  & $H_\textit{DVF}$     & 1.000 (203)  & 1.000 (279)      & 0.088 (342)      & 0.955 (301)          & 0.363 (327)          & -                & $0.000^+$ (444)   \\
			                                  & $H_\textit{RAND}$    & 1.000 (116)  & 1.000 (155)      & 1.000 (194)      & 1.000 (184)          & 1.000 (196)          & 1.000 (196)      & -                 \\
			\midrule
			\multirow{7}{*}{$f_\textit{RAI}$} & Our approach         & -            & $0.000^+$ (414)  & $0.000^+$ (429)  & $0.000^+$ (406)      & $0.000^+$ (421)      & $0.000^+$ (420)  & $0.000^+$ (518)   \\
			                                  & $H_\textit{PWI}$     & 1.000 (226)  & -                & $0.000^+$ (344)  & 0.024 (333)          & $0.000^+$ (347)      & $0.000^+$ (362)  & $0.000^+$ (481)   \\
			                                  & $H_\textit{RAI}$     & 1.000 (211)  & 1.000 (296)      & -                & 0.994 (308)          & 0.855 (289)          & 0.871 (304)      & $0.000^+$ (446)   \\
			                                  & $H_{\textit{PWI}-2}$ & 1.000 (234)  & 0.976 (307)      & 0.006 (332)      & -                    & 0.018 (346)          & 0.097 (333)      & $0.000^+$ (452)   \\
			                                  & $H_{\textit{RAI}-2}$ & 1.000 (219)  & 1.000 (293)      & 0.145 (351)      & 0.982 (294)          & -                    & 0.622 (312)      & $0.000^+$ (437)   \\
			                                  & $H_\textit{DVF}$     & 1.000 (217)  & 1.000 (278)      & 0.129 (336)      & 0.903 (307)          & 0.378 (328)          & -                & $0.000^+$ (442)   \\
			                                  & $H_\textit{RAND}$    & 1.000 (122)  & 1.000 (159)      & 1.000 (194)      & 1.000 (188)          & 1.000 (203)          & 1.000 (198)      & -                 \\
			\bottomrule
		\end{tabular}
	}
\end{table}

	\section{Visualized Results of the Computational Experiments on the Questioning Policy} \label{app-J}
	This section provide a detailed visualization of the performance of the proposed questioning policy and baseline policies under varying numbers of alternatives, criteria, and interaction rounds.

	\subsection{Performance analysis for different numbers of alternatives and criteria}

	Figures \ref{fig:metrics_alternatives} and \ref{fig:metrics_criteria} present the results of the three metrics $f_\textit{VAR}$, $f_\textit{PWI}$, and $f_\textit{RAI}$, characterizing the uncertainty of the DM's preferences after $s$ interactions guided by the different questioning policies, for problems involving $n$ alternatives and $3$ criteria and problems involving $8$ alternatives and $m$~criteria respectively, where $s \in \{2, 4, 6, 8\}$, $n \in \{6, 7, 8, 9, 10\}$, $m \in \{2, 3, 4, 5\}$. We observe that the proposed MCTS-based questioning policy consistently outperforms the other comparative methods across most problem settings and evaluation metrics, whereas the random strategy consistently performs the worst across all problem settings.

	The performances of these seven methods have the same trend with the variation of the problem settings. On the one hand, as the number of alternatives increases, $f_\textit{RAI}$ ascends for all the methods, whereas the variations in $f_\textit{VAR}$ and $f_\textit{PWI}$ do not exhibit a clear pattern. This is because $f_\textit{RAI}$ quantifies the uncertainty in the ranks assigned to all the alternatives; more alternatives incur more positions in the ranking, leading to greater uncertainty in terms of $f_\textit{RAI}$. 
	By contrast, $f_\textit{VAR}$ and $f_\textit{PWI}$ quantify the variance of the posterior distribution derived through the proposed variational Bayesian approach and the uncertainty of the preference relations for all pairs of alternatives, respectively. Thus, they are not easily affected by changes in the number of alternatives. On the other hand, as the number of criteria increases, the performance of the seven questioning policies in terms of the three metrics does not exhibit a clear trend.

	\begin{figure}[!htbp]
		\centering
		\includegraphics[scale=0.45]{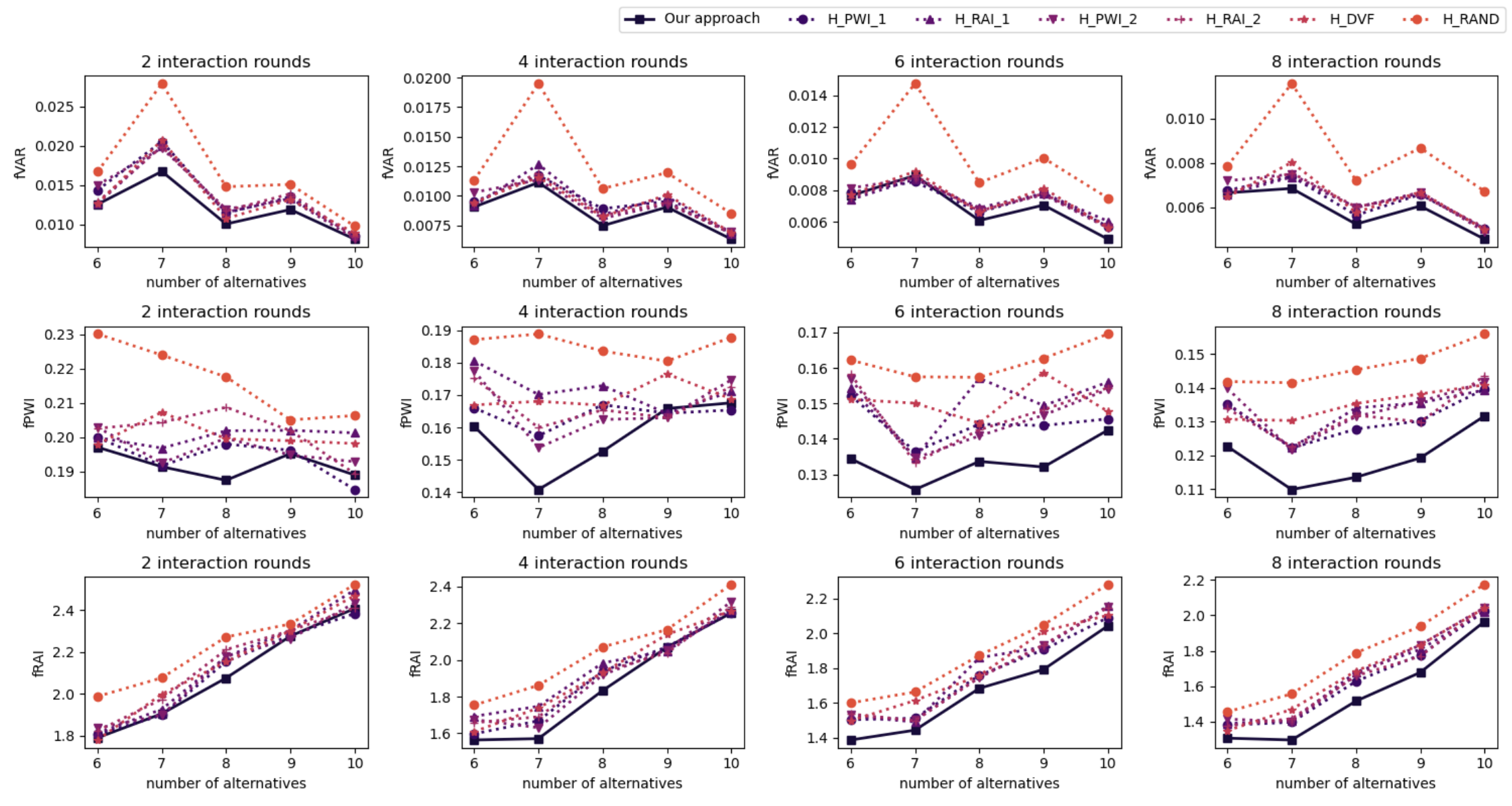}
		\caption{Average values of $f_\textit{VAR}$, $f_\textit{PWI}$, and $f_\textit{RAI}$ after 2, 4, 6 and 8 interaction rounds with 7 questioning policies for the test problems with 6-10 alternatives and 3 criteria.} \label{fig:metrics_alternatives}
	\end{figure}
	\begin{figure}[!htbp]
		\centering
		\includegraphics[scale=0.45]{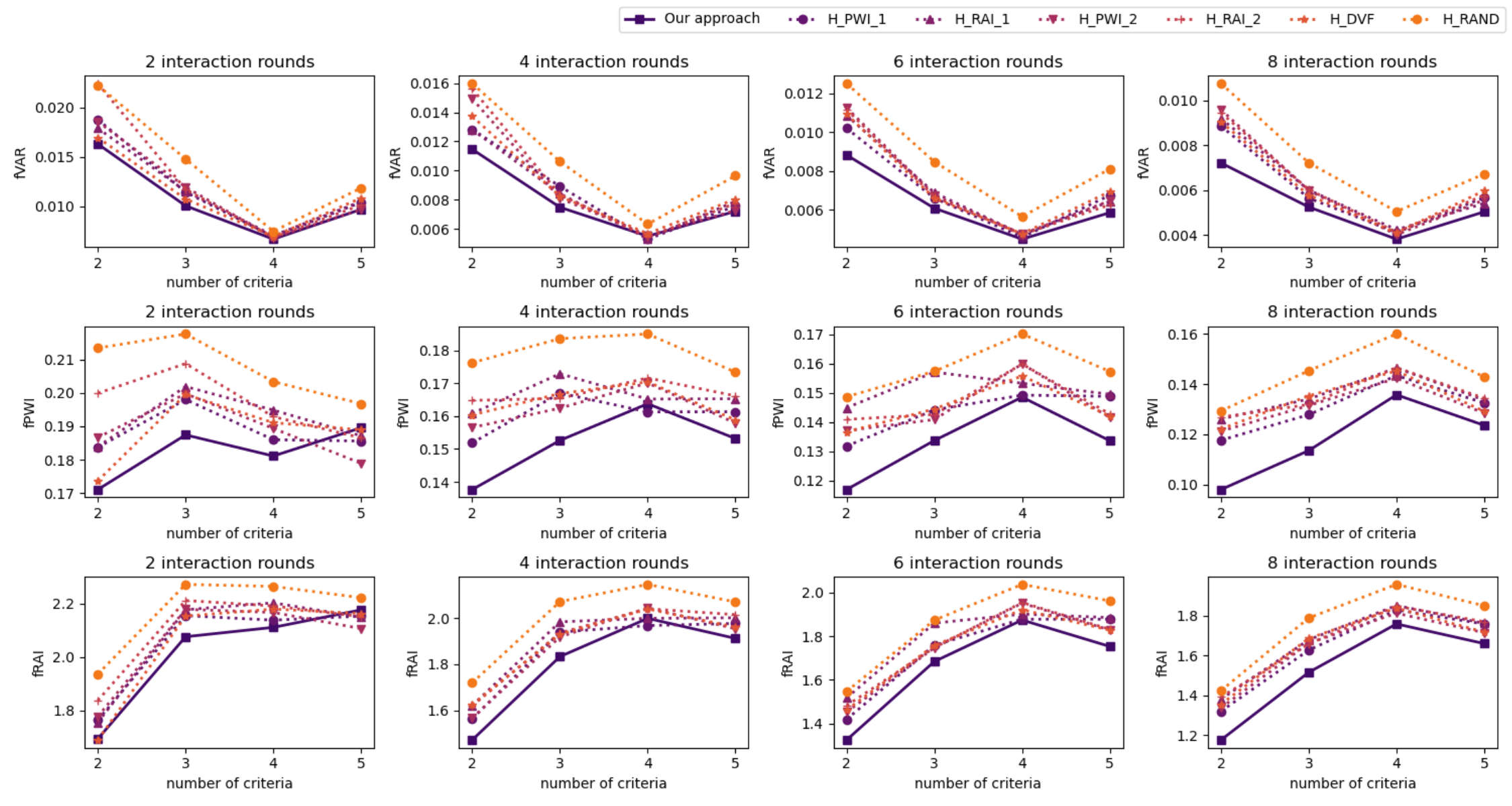}
		\caption{Average values of $f_\textit{VAR}$, $f_\textit{PWI}$, and $f_\textit{RAI}$ after 2, 4, 6 and 8 interaction rounds with 7 questioning policies for the test problems with 2-5 criteria and 8 alternatives.} \label{fig:metrics_criteria}
	\end{figure}

	\subsection{Performance analysis for different numbers of interaction rounds} 

	Figure \ref{fig:metrics_Questions} presents the average values of $f_\textit{VAR}$, $f_\textit{PWI}$, and $f_\textit{RAI}$ after 2, 4, 6, and 8 interactions with 7 questioning policies for test problems with different numbers of alternatives and criteria. This provides an intuitive understanding of how the number of interaction rounds impacts the results from another perspective. The observation is two-fold. On the one hand, as the number of interaction rounds increases, the results for all methods show a decline across all metrics. This aligns with the fact that more preference information reduces our uncertainty about the DM's preferences. On the other hand, with more interactions, the advantages of our method become more pronounced in most problem settings. This is because our method employs a decision-time planning approach, which allocates the limited computational budget to rollouts that simulate sequences starting from the current state, allowing for careful consideration of the long-term impacts of decisions while still focusing on the current state. As a result, it mitigates the shortsightedness issue found in comparative methods, with this advantage becoming more evident as the number of interaction rounds increases.

	\begin{figure}[!h]
		\centering
		\includegraphics[scale=0.45]{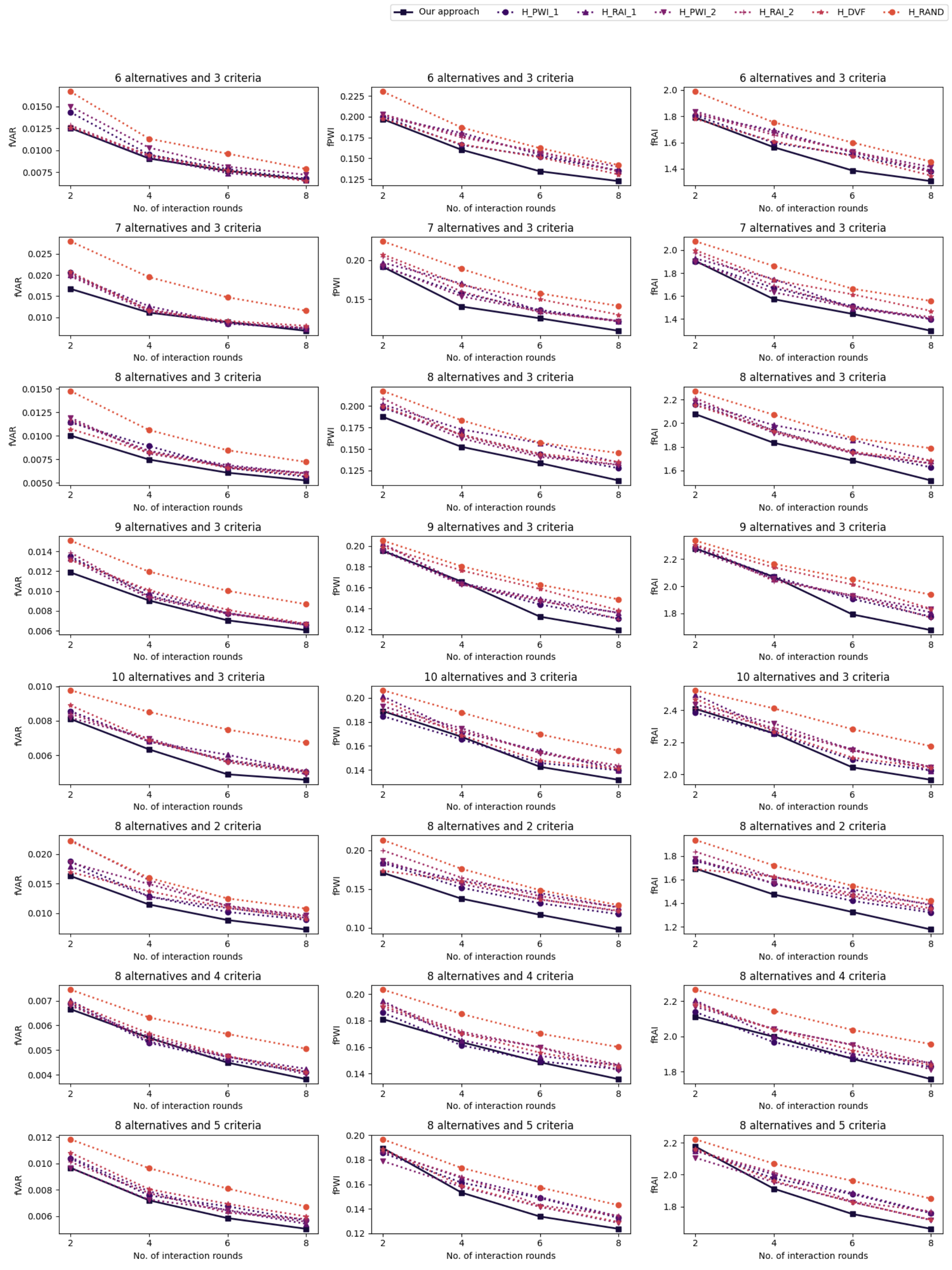}
		\caption{Average values of $f_\textit{VAR}$, $f_\textit{PWI}$, and $f_\textit{RAI}$ after 2, 4, 6 and 8 interaction rounds with 7 questioning policies for the test problems with different numbers of alternatives and criteria.} \label{fig:metrics_Questions}
	\end{figure}

\end{APPENDICES}
  
  \end{document}